\documentclass[lettersize,journal]{IEEEtran}
\usepackage{amsmath,amsfonts}
\usepackage{algorithmic}
\usepackage{algorithm}
\usepackage{array}
\usepackage[caption=false,font=normalsize,labelfont=sf,textfont=sf]{subfig}
\usepackage{textcomp}
\usepackage{stfloats}
\usepackage{url}
\usepackage{verbatim}
\usepackage{graphicx}
\usepackage{cite}

\usepackage{xcolor}
\usepackage{color}
\usepackage{enumerate}
\usepackage{arydshln}
\usepackage{tikz}
\usepackage{multirow}
\usepackage{microtype}
\usepackage{booktabs} 
\usepackage{threeparttable, adjustbox}
\usepackage{float}
\usepackage{amssymb}
\usepackage{amsmath}
\usepackage{lscape}
\usepackage{dsfont}

\begin{document}

\title{WavLM: Large-Scale Self-Supervised  Pre-Training  \\ for Full Stack Speech Processing}

\author{Sanyuan Chen*, Chengyi Wang*, Zhengyang Chen*, Yu Wu*, Shujie Liu, Zhuo Chen, \\ Jinyu Li, Naoyuki Kanda, Takuya Yoshioka, Xiong Xiao, Jian Wu, Long Zhou, Shuo Ren, \\ Yanmin Qian, Yao Qian, Jian Wu, Michael Zeng,  Xiangzhan Yu, Furu Wei
\thanks{* Equal contribution. The work was done at Microsoft during the internship of the first three authors.}
\thanks{Correspondence to Yu Wu: yuwu1@microsoft.com}
}

\maketitle

\begin{abstract}
Self-supervised learning (SSL) achieves great success in speech recognition, while limited exploration has been attempted for other speech processing tasks. As speech signal contains multi-faceted information including speaker identity, paralinguistics, spoken content, etc., learning universal representations for all speech tasks is challenging. To tackle the problem, we propose a new pre-trained model, WavLM, to solve full-stack downstream speech tasks. WavLM jointly learns masked speech prediction and denoising in pre-training. By this means, WavLM does not only keep the speech content modeling capability by the masked speech prediction, but also improves the potential to non-ASR tasks by the speech denoising. In addition, WavLM employs gated relative position bias for the Transformer structure to better capture the sequence ordering of input speech. We also scale up the training dataset from 60k hours to 94k hours. WavLM Large achieves state-of-the-art performance on the SUPERB benchmark, and brings significant improvements for various speech processing tasks on their representative benchmarks. The code and pre-trained models are available at \url{https://aka.ms/wavlm}.

\end{abstract}


\section{Introduction}
Over the past few years, self-supervised learning (SSL) has achieved great success in the fields of natural language processing (NLP) \cite{bert,xlnet,t5}. It leverages large amounts of text data to learn universal text representations, which can benefit almost all NLP downstream tasks by fine-tuning. Recently, SSL has also shown prominent results for speech processing, especially on phoneme classification \cite{cpc} and automatic speech recognition (ASR) \cite{wav2vec2,hsu2021hubert,wang2021unispeech}. However, in other speech tasks, it is still the standard practice to train models from scratch with task-specific datasets. 

Building a general pre-trained model for full stack speech processing tasks is essential to the further development of speech processing, because many tasks are short of supervised data, especially for non-ASR tasks. A model pre-trained on large-scale unlabeled data is able to boost the performance of these tasks, reduce data labeling efforts, and lower entry barriers for individual tasks. Furthermore, it is infeasible to build different pre-trained models for different downstream tasks, as the pre-training stage requires huge computational resources. 
In the past, it has been infeasible to build such a general model, as different tasks focus on different aspects of speech signals. For instance,  speaker verification requires the network to learn the speaker characteristic regardless of the spoken content, while speech recognition demands the network to discard speaker characteristics and focus only on the content information.  Meanwhile, unlike verification and recognition tasks,  speaker diarization and speech separation involve multiple speakers, which creates additional obstacles to learning general speech representations. Recent advances fueled by large-scale pre-trained models have changed the situation. \cite{superb} proves the potential of pre-trained models on full-stack speech tasks by using the weighted sum of embeddings from different layers.\footnote{The paper does not explicitly mention it, but their presentation highlights the contribution of weighted sum hidden states. Details can be found from 28:00 to 31:00 of \url{https://www.youtube.com/watch?v=Fw2ujGzmfNA}} They find different layers containing information useful for different tasks. For instance, the hidden states of the top layers are useful for ASR, while the bottom layers are more effective for speaker verification. 

While exciting as a proof of concept, there are still some drawbacks in existing pre-trained models: 1) Current pre-trained models are unsatisfactory for multi-speaker tasks, such as speaker diarization and speech separation. Our experiments show that speech separation models trained on top of HuBERT \cite{hsu2021hubert}, a top-performing speech pre-trained model, achieve only marginal improvement compared with the models trained from scratch. This is mainly because the pre-training methods do not sufficiently enforce speaker discrimination, and the training data contain only single-speaker audios. 
2) Speech pre-training crucially relies on high quality and large quantities of unlabeled audios. The existing system utilizes Libri-Light   \cite{librilight} as the main source, but the massive audiobook data mismatches the data in a real scenario and using it exclusively hurts the model performance when the acoustic characteristics of the downstream tasks are different from those of the audiobook \cite{chan2021speechstew,likhomanenko2020rethinking,narayanan2018toward,narayanan2019recognizing}.  \cite{hsu2021robust}  trains wav2vec 2.0  \cite{wav2vec2} on larger and more diverse datasets, but there are still over 90$\%$ audio data derived from audiobook. To eliminate the audiobook data bias, we try to gather data from different sources as much  as possible in our experiments. 

In this paper, we present WavLM, which learns universal speech representations from massive unlabeled speech data and adapts effectively across various speech processing tasks. We propose a masked speech denoising and prediction framework for WavLM, where some inputs are simulated noisy/overlapped speech with masks and the target is to predict the pseudo-label of the original speech on the masked region like HuBERT. The framework combines the masked speech prediction and denoising in pre-training. Therefore, the WavLM model learns not only the ASR information by the masked speech prediction, but also the knowledge of non-ASR tasks by the speech denoising modeling. For instance, the process of pseudo-label prediction on overlapped speech implicitly improves the model capability on diariazation and separation tasks. The speaker identity information and speech enhancement capability are modeled by the pseudo-label prediction on simulated noisy speech.

In addition, we optimize the model structure and training data of HuBERT and wav2vec 2.0.  
We add gated relative position bias (grep) \cite{xlm-e} to the Transformer structure as the backbone, which improves model performance for ASR and keeps almost the same parameter number and training speed. Compared with the convolutional relative position embedding used in wav2vec 2.0 and HuBERT, the gates allow the relative position bias to be adjusted adaptively by conditioning on the current speech content. 
To further improve the model robustness and alleviate the data mismatch, we scale up unlabeled pre-training data to 94k hours of public audios. The dataset consists of 60k hours of Libri-Light, 10k hours of GigaSpeech \cite{GigaSpeech2021}, and 24k hours of VoxPopuli \cite{wang2021voxpopuli}. The new dataset consists of training instances from different scenarios, such as podcasts, YouTube, and European Parliament (EP) event recordings.

We evaluate our models on \textbf{nineteen} subtasks, fifteen of which are from SUPERB, and the other four are classic speech tasks on their representative testsets.

\begin{itemize} \item 
WavLM achieves state-of-the-art (SOTA) performance on \textbf{SUPERB} \cite{superb,tsai2022superb}. WavLM Large outperforms HuBERT Large on \textbf{14} subtasks, and achieves an absolute \textbf{2.4} point improvement in the overall evaluation. Even WavLM Base+, a 3 times smaller model, is better than HuBERT Large owing to our three modifications. 

\item \textbf{Speaker verification} is a task to verify the speaker's identity from the voice characteristics. We select this task to evaluate the model's capability of extracting speaker-related features.  
WavLM Large exceeds the well-known SOTA system, ECAPA-TDNN \cite{desplanques2020ecapa}, by a large margin  and achieves \textbf{0.383}\%, \textbf{0.480}\% and \textbf{0.986}\% EER (Equal Error Rate) on the three official trial lists of VoxCeleb1 \cite{nagrani2017voxceleb}.

\item \textbf{Speech separation} is a classic multi-speaker task, which is the key to solving the cocktail party problem. The task can evaluate the model's capability of extracting multiple speech signals from a mixture of sounds. WavLM achieves SOTA performance on the speech separation LibriCSS benchmark \cite{chen2020continuous}, and significantly outperforms the previous Conformer model \cite{chen2020continuous2} by a  \textbf{27.7}$\%$ relative word error rate (WER) reduction.

\item \textbf{Speaker diarization}  is a task to recognize ``who spoke when'' from an input audio stream \cite{park2021review}.  WavLM achieves SOTA performance on the CALLHOME speaker diarization benchmark. Compared to the EEND-EDA clustering method \cite{horiguchi2021towards}, our model achieves a  \textbf{12.6}$\%$ diarization error rate reduction. 

\item \textbf{Speech recognition} requires the model to learn content information, which is the main focus of the previous SSL work. We evaluate our model in the LibriSpeech 960h setting.   WavLM shows comparable performance to the wav2vec 2.0 and HuBERT, which achieves 1.8\% and 3.2\% WER  on the test-clean and test-other sets, respectively. 

\end{itemize}
The contribution of the paper can be summarized as follows:
1) WavLM sheds light on a general pre-trained model for full stack speech processing tasks, in contrast to the previous SSL works focusing on a group of  similar tasks.
2) We propose simple but effective modifications to the existing pre-trained models, which show general and consistent improvements across downstream tasks.
3) We scale-up self-supervised speech pre-training with more unlabeled data and longer training steps. 
4) We achieve SOTA results on the SUPERB benchmark, and  significantly boost the performance  for various speech processing tasks on their representative benchmarks, including speech separation, speaker verification, and speaker diarization. The models and code are released \footnote{\url{https://aka.ms/wavlm}} to facilitate future research.

\section{Related Work}

 SSL methods can be categorized into generative learning, discriminative learning, and multitask learning, based on the training objective. The research line of {generative learning} can be traced back to the auto-encoding model, which reconstructs the whole speech from latent variables, either continuous \cite{Audio_Word2Vec,hsu2017learning,hsu2017learning2} or discrete \cite{chorowski2019unsupervised}. Recent works propose to  predict future frames from the history with an autoregressive model \cite{vq_apc,apc1,apc2,improved_apc}, or recover the masked frames from the corrupted speech with a non-autoregressive model \cite{npc,tera,mockingjay,decoar,decoar2,gong2021ssast}. Apart from generative learning, discriminative learning has also gathered interests recently. The well-known examples include CPC \cite{cpc}, wav2vec \cite{wav2vec}, vq-wav2vec \cite{vq_wav2vec}, wav2vec 2.0 \cite{wav2vec2}, DiscreteBERT \cite{vq_wav2vec_ft}, HuBERT \cite{hsu2021hubert} and w2v-BERT \cite{chung2021w2v}. CPC and the wav2vec series models use the contrastive InfoNCE loss to discriminate the positive samples from negative samples. Motivated by the masked language model loss in NLP, DiscreteBERT and HuBERT predict discrete targets of masked regions. 
 w2v-BERT further combines the contrastive
loss and the masked prediction loss in an end-to-end fashion.
{Multi-task learning} is adopted in PASE \cite{pase} and PASE+ \cite{pase+}. They employ lots of pre-training objectives such as waveform generation, prosody regression, and contrastive objectives. UniSpeech \cite{wang2021unispeech} and JUST \cite{bai2022joint} combine SSL and supervised learning for ASR, and show impressive results on multi-lingual test sets. 

Unlike SSL in computer vision (CV) and NLP fields, where one pre-trained model is adapted to various downstream tasks, most speech SSL methods focus on phoneme classification and ASR. Recently, \cite{superb} proposed the SUPERB benchmark to evaluate SSL models across different tasks. According to the results, HuBERT enjoys the best generalization ability in the overall evaluation. To better learn speaker characteristics, \cite{chen2021unispeech} proposed UniSpeech-SAT, which extends the HuBERT framework with speaker-aware pre-training. It significantly outperforms other pre-trained models on the speaker-related tasks with a slight degradation on the ASR.

Compared with existing works, our model is the first to explore SSL for full stack tasks instead of focusing on ASR or other specific tasks. It should be noted that a concurrent work BigSSL \cite{zhang2021bigssl} also mentions large SSL model could handle various speech tasks. The difference is that our work demonstrates that the full stack tasks can be handled by the careful pre-training and fine-tuning strategy design, even without scaling up the model size to 8 billion parameters.

\section{Background: HuBERT}
HuBERT is an SSL method that benefits from an offline clustering step to provide target labels for a BERT-like prediction loss \cite{bert}. The backbone is a Transformer encoder \cite{vaswani2017attention} with $L$ blocks. During pre-training, the Transformer consumes masked acoustic features $\mathbf{u}$ and outputs hidden states $\mathbf{h}^L$. The network is optimized to predict the discrete target sequence $\mathbf{z}$, where each $z_t \in [C]$ is a $C$-class categorical variable. The distribution over codewords is parameterized with 
\begin{equation}
    p(c|\mathbf{h}_t^L) = \frac{{\rm exp}({\rm sim}(\mathbf{h}_t^L\mathbf{W}^P, \mathbf{e}_c)/\tau)}{\sum_{c'=1}^{C}{\rm exp}({\rm sim}(\mathbf{h}_t^L\mathbf{W}^P, \mathbf{e}_{c'})/\tau)}
\label{eq1}
\end{equation} 
where $\mathbf{W}^P$ is a projection matrix, $\mathbf{h}_t^L$ is the output hidden state for step $t$, $\mathbf{e}_c$ is the embedding for codeword $c$, sim($a, b$) computes the cosine similarity and $\tau=0.1$ scales the logit. HuBERT proposes a masked speech prediction task, where the prediction loss is only applied over the masked regions, forcing the model to learn a combined acoustic and language model over the continuous inputs.

HuBERT adopts an iterative re-clustering and re-training process: For the first iteration, the targets are assigned by clustering the MFCC features of the training data; For the second iteration, a new generation of training targets are created by clustering the latent representations generated by the first iteration trained model. 
\label{ss:hubert}

\begin{figure}
\centering
  \includegraphics[width=0.48\textwidth]{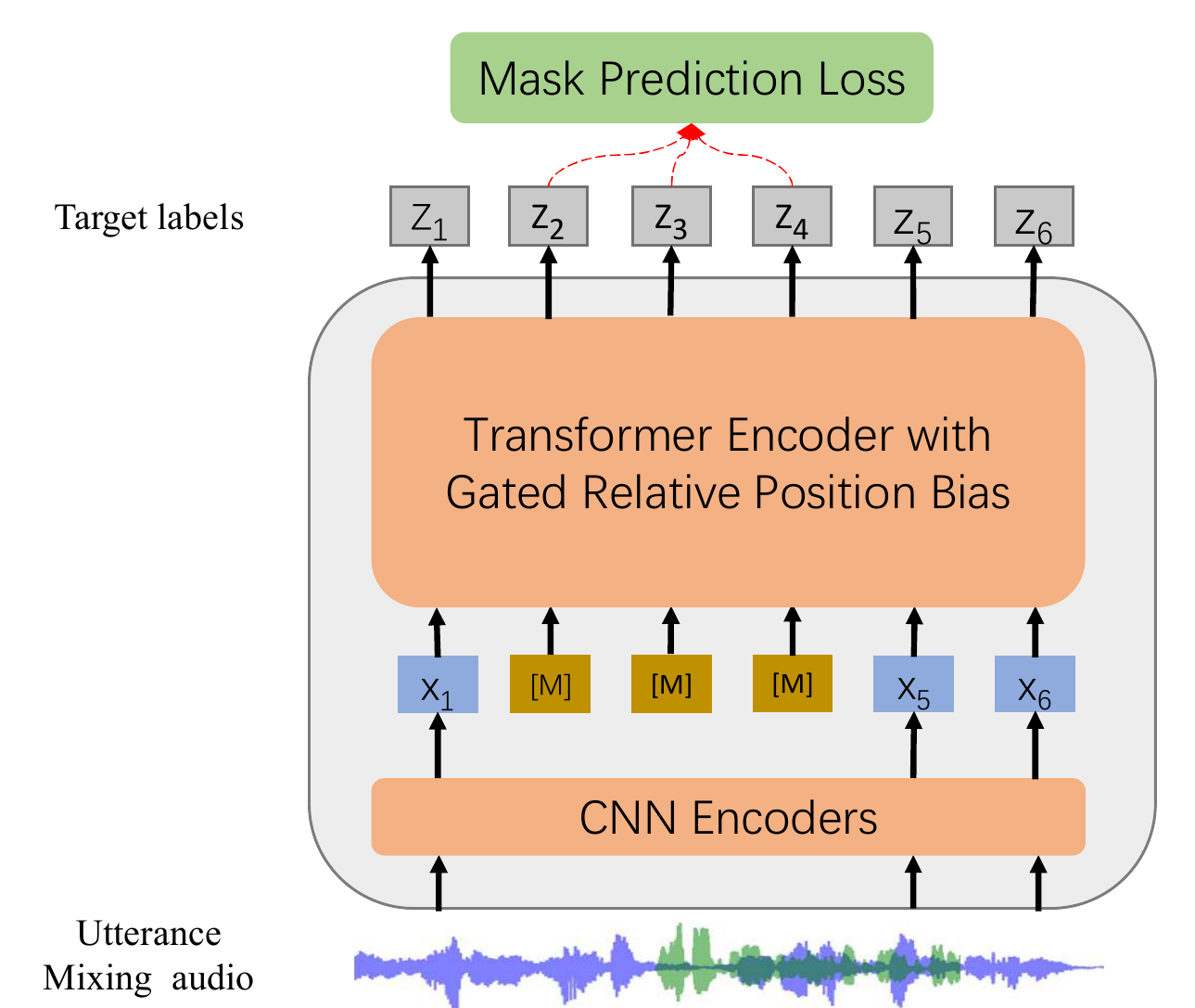}
  \caption{Model Architecture.}
\label{model}
\end{figure}

\section{WavLM}
We propose a masked speech denoising and prediction framework, where some inputs are simulated noisy/overlapped with masks and the target is to predict pseudo-labels of the original speech on the masked region. 
Unlike existing masked speech modeling (HuBERT), which just focuses on the ASR task, the masked speech denoising allows us to extend pre-trained speech models to non-ASR tasks, since it implicitly models information we need in the speaker identification, separation, and diarization tasks.  We further optimize the Transformer backbone and extend pre-training data to 94k public English data. 
\subsection{Model Structure}
\label{sec:grep}
Our model architecture uses the Transformer model as the backbone. As shown in Figure \ref{model}, it contains a convolutional feature encoder and a Transformer encoder. The convolutional encoder is composed of seven blocks of temporal convolution followed by layer normalization and a GELU activation layer. The temporal convolutions have 512 channels with strides (5,2,2,2,2,2,2) and kernel widths (10,3,3,3,3,2,2), resulting in each output representing about 25ms of audio strode by 20ms. The convolutional output representation $\mathbf{x}$ is masked as the Transformer input. The Transformer is equipped with a convolution-based relative position embedding layer with 128 kernel size and 16 groups at the bottom. 

 To improve the model, we employ gated relative position bias \cite{xlm-e} which is encoded based on the offset between the ``key" and ``query" in the Transformer self-attention mechanism. Let $\{\mathbf{h}_i\}_{i=1}^{T}$ denote the input hidden states for the self-attention module, each $\mathbf{h}_i$ is linearly projected to a triple of query, key and value $(\mathbf{q}_i, \mathbf{k}_i, \mathbf{v_i})$ as:
\begin{equation}
    \mathbf{q}_i,\mathbf{k}_i,\mathbf{v}_i = \mathbf{h}_i \mathbf{W}^Q, \mathbf{h}_i \mathbf{W}^K, \mathbf{h}_i \mathbf{W}^V
\end{equation}
The self-attention outputs $\{\tilde{\mathbf{h}}_i\}_{i=1}^{T}$ are computed via:
\begin{align}
a_{ij} &\propto \exp\{ \frac{ \mathbf{q}_i \cdot \mathbf{k}_j }{\sqrt{d_{k}}} + r_{i-j} \}\label{eq3} \\
\tilde{\mathbf{h}}_i &= \sum_{j=1}^{T}{ a_{ij} \mathbf{v}_j } 
\end{align}
where $r_{i-j}$ is the gated relative position bias added to the attention logits. It is computed by:
\begin{align}
g_i^\text{(update)} , g_i^\text{(reset)} &= \sigma ( \mathbf{q}_i \cdot \mathbf{u} ), \sigma ( \mathbf{q}_i \cdot \mathbf{w} ) \nonumber \\
\tilde{r}_{i-j} &= w g_i^\text{(reset)} d_{i-j} \nonumber \\
r_{i-j} = d_{i-j} + g_i^\text{(update)} &d_{i-j} + (1 - g_i^\text{(update)}) \tilde{r}_{i-j} \nonumber
\end{align}
where $d_{i-j}$ is a learnable scalar relative position bias, the vectors $\mathbf{u},\mathbf{w} \in \mathbb{R}^{d_k}$ are learnable parameters, $\sigma$ is a sigmoid function, and $w$ is a learnable value.

In our work, $d_{i-j}$ is a bucket relative position embedding~\cite{t5} and the embedding parameters are shared across all layers. We use $n=320$ embeddings and each corresponds to a range of possible $(i-j)$ offsets. The range increased logarithmically up to a maximum offset of $m=800$, beyond which we assign all relative offsets to the same embedding, i.e.,

\begin{equation}
d_{|i-j|}=
\begin{cases}
|i-j|,  & |i-j|< \frac{n}{4} \\
 \lfloor \frac{n}{4} (\frac{\log (|i-j| ) - \log (\frac{n}{4})}{\log(m) - \log (\frac{n}{4})} + 1)  \rfloor, & \frac{n}{4} \le |i-j| < m \\
\frac{n}{2} - 1, & |i-j|\ge m \\
\end{cases} 
\end{equation}
\begin{equation}
d_{i-j}=d_{|i-j|} + \frac{n}{2} \cdot \mathds{1}_{\{i-j>0\}} 
\end{equation}

Compared with the convolutional relative position embedding in wav2vec 2.0 and HuBERT, the gates take the content into consideration, and adaptively adjust the relative position bias by conditioning on the current speech content. Intuitively, the same distance offset between two frames tends to play different roles if one frame is the silence while the other belongs to a speech segment.

\begin{algorithm*}[t]
\caption{Noisy/Overlapped Speech Simulation}
\footnotesize
\label{algo:utt_mixing}
\begin{algorithmic}[1]
\STATE{\textbf{given} a batch of speech utterances $\mathbf{U} = \{\mathbf{u}^i\}_{i=1}^B$ with batch size $B$ and length $L$, mixing probability $p$}, a set of DNS noises $\mathbf{N} = \{\mathbf{n}^i\}_{i=1}^M$ with size $M$, mixing noise probability $p_n$
\STATE{Choose $S$ utterances $\mathbf{U}^S \subset \mathbf{U}$ by Bernoulli sampling with probability $p$}
\FOR{each primary utterance $\mathbf{u}^{\text{pri}} \in \mathbf{U}^S$}
        \STATE{Sample a random value $v$ from the continuous uniform distribution $\mathcal{U}(0, 1)$}
        \IF{$v > p_n$}
        \STATE{Sample a secondary utterance $\mathbf{u}^{\text{sec}}$ from discrete uniform distribution with probability $P(\mathbf{u}^{\text{sec}} = \mathbf{x}) = \frac{1}{B}, \mathbf{x} \in  \mathbf{U}$}
        \STATE{Sample the mixing energy ratio $r$ from the continuous uniform distribution $\mathcal{U}(-5, 5)$ }
        \ELSE
        \STATE{Sample a noise $\mathbf{u}^{\text{sec}}$ from discrete uniform distribution with probability $P(\mathbf{u}^{\text{sec}} = \mathbf{x}) = \frac{1}{M}, \mathbf{x} \in  \mathbf{N}$}
        \STATE{Sample the mixing energy ratio $r$ from the continuous uniform distribution $\mathcal{U}(-5, 20)$ }
        \ENDIF
        \STATE{Sample the mix length $l$ from discrete uniform distribution with probability $P(l = x) = \frac{2}{L}, x \in \{1, \cdots, \frac{L}{2}\}$}
        \STATE{Sample a start position $s^{\text{pri}}$ of $\mathbf{u}^{\text{pri}}$ from discrete uniform distribution with probability $P(s^{\text{pri}} = x) = \frac{1}{L - l}, x \in \{1, \cdots, L - l\}$}
        \STATE{Sample a start position $s^{\text{sec}}$ of $\mathbf{u}^{\text{sec}}$ from discrete uniform distribution with probability $P(s^{\text{sec}} = x) = \frac{1}{L - l}, x \in \{1, \cdots, L - l\}$}
        \STATE{Calculate the energy of the primary utterance $E^{\text{pri}} \leftarrow \frac{\sum \mathbf{u}^{\text{pri}} \cdot \mathbf{u}^{\text{pri}}}{L}  $ }
        \STATE{Calculate the energy of the secondary utterance $E^{\text{sec}} \leftarrow \frac{\sum \mathbf{u}^{\text{sec}} \cdot \mathbf{u}^{\text{sec}}}{L} $ }
        \STATE{Calculate the mixing scale $scl \leftarrow \sqrt{\frac{E^{\text{pri}}}{10^{\frac{r}{10}}E^{\text{sec}}}} $ }
        \STATE{$\mathbf{u}^{\text{pri}}[s^{\text{pri}}: s^{\text{pri}}+l] \leftarrow \mathbf{u}^{\text{pri}}[s^{\text{pri}}: s^{\text{pri}}+l] + scl \cdot \mathbf{u}^{\text{sec}}[s^{\text{sec}}: s^{\text{sec}}+l] $}
\ENDFOR
\STATE{\textbf{return} $\mathbf{U}$}
\end{algorithmic}
\end{algorithm*}
\raggedbottom

\subsection{Masked Speech Denoising and Prediction}

We propose a masked speech denoising and prediction framework to improve model robustness for complex acoustic environments and the preservation of speaker identity. Specifically, we manually simulated noisy/overlapped speech as inputs, and predict the pseudo-labels of original speech on the masked region. 

We simulate the noisy speech with multiple speakers and various background noise for self-supervised pre-training. We randomly select some utterances from each training batch and mix them with a randomly selected noise audio or secondary utterance at a random region. The noise audio and secondary utterance are randomly selected from the same batch, randomly cropped, and scaled by a random source energy ratio. We ensure that the overlap region is less than 50\% and take the speaker from the first utterance as the main speaker.
With the masked speech denoising and prediction task, the model is trained to identify the main speaker from the noisy/overlapped speech and predict the content information corresponding to the main speaker with the mask prediction loss.
\subsubsection{Noisy/Overlapped Speech Simulation}

The details of our noisy/overlapped speech simulation method are shown in Algorithm~\ref{algo:utt_mixing}.
Given a batch of speech utterances $\mathbf{U} = \{\mathbf{u}^i\}_{i=1}^B$ with batch size $B$ and length $L$ and a set of DNS (Deep Noise Suppression) noises \cite{reddy2021interspeech} $\mathbf{N} = \{\mathbf{n}^i\}_{i=1}^M$ with size $M$ (line 1), we first randomly choose $S$ utterances to mix $\mathbf{U}^S = \{\mathbf{u}^i\}_{i=1}^S$ from the batch by Bernoulli sampling with probability $p$ (line 2). 
Then, for each utterance $\mathbf{u}^{\text{pri}} \in \mathbf{U}^S$ (line 3), 
we sample a random value $v$ the continuous uniform distribution $\mathcal{U}(0, 1)$ to decide whether to mix noise or a secondary utterance  (line 4). 
If the random value $v$ is greater than the mixing noise probability $p_n$ (line 5), we sample a secondary  utterance $\mathbf{u}^{\text{sec}}$ from a discrete uniform distribution over the batch $\mathbf{U}$ (line 6), and randomly sample the mixing energy ratio $r$ from the uniform distribution $\mathcal{U}(-5, 5)$ (line 7).
Otherwise, we sample a noise $\mathbf{u}^{\text{sec}}$ from a discrete uniform distribution over the set of DNS noises $\mathbf{N}$ (line 9), and randomly sample the mixing energy ratio $r$ from the uniform distribution $\mathcal{U}(-5, 20)$ (line 10).
The sample range of the mixing energy ratio follows the typical training utterance simulation process of speech separation task \cite{wu21f_investigation}.
Then, we randomly select the mixing regions for both the utterances from the uniform distributions (line 12-14).
The mixing length $l$ is uniformly sampled from  $\{1, \cdots, \frac{L}{2}\}$ (line 12), and the start positions $s^{\text{pri}}$ and $s^{\text{sec}}$ for utterance $\mathbf{u}^{\text{pri}}$ and $\mathbf{u}^{\text{sec}}$ are both uniformly sampled from $\{1, \cdots, L - l\}$ (line 13 and 14).
Note  that as the  mixing  portion  in  each  utterance is constrained  to  be  less  than  50\%, 
the primary utterance would always be longer than the secondary utterance, avoiding the problem of the indistinguishable main speaker in the mixed speech signals.
Next, given the mixing regions of the primary utterance $\mathbf{u}^{\text{pri}}[s^{\text{pri}}: s^{\text{pri}}+l]$ and the secondary utterance $\mathbf{u}^{\text{sec}}[s^{\text{sec}}: s^{\text{sec}}+l]$, we calculate the corresponding mixing scale of the secondary utterance $scl$ with the energy of the primary utterance $E^{\text{pri}}$, the secondary utterance $E^{\text{sec}}$ (line 15-17).
Finally, we mix the selected region of the primary utterance $\mathbf{u}^{\text{pri}}[s^{\text{pri}}: s^{\text{pri}}+l]$ with  the secondary utterance $\mathbf{u}^{\text{sec}}[s^{\text{sec}}: s^{\text{sec}}+l]$ scaled by the mixing scale $scl$ (line 18).

\subsubsection{Mask Prediction Loss}

Following HuBERT, we use the mask prediction loss to optimize our network. Suppose that we have an utterance $\mathbf{u}$ and its simulated version  $\mathbf{u'}$, we always generate pseudo-labels $\mathbf{z}$ by feeding  $\mathbf{u}$ to the last iteration network. We follow HuBERT using the k-means clustering center on MFCC or latent representations as the pseudo-labels. The details will be introduced in Section \ref{setup}. Then, we obtain the hidden state $\mathbf{h}_t^L$ by feeding $\mathbf{u'}$ to the current network, and optimize the objective function:
\begin{equation}
\label{eq:mploss}
   \mathcal{ L}  = -\sum_{l \in K}\sum_{t \in M}{\rm log}p(z_t|\mathbf{h}_t^L) 
\end{equation}
where $M$ denotes the set of masked indices in time domain and $\mathbf{h}_t^L$ is the $L$-layer Transformer output for step $t$. 
Compared to previous methods, the framework is more beneficial to various non-ASR tasks, since it implicitly models the non-ASR information in pre-training. 

\subsection{Pre-Training Data }
We leverage large-scale unsupervised data from diverse domains to improve the robustness of our model. 
Previous works use LibriSpeech \cite{librispeech} or LibriLight \cite{librilight} datasets for pre-training, which limits the generalization capability of the pre-trained model since the input data are all extracted from the audiobook.  The background acoustics of the speech obtained from the audiobook is different from what is observed in 
other conditions
, since the real captured sounds are usually accompanied by various types of noise.

Motivated by this, we extend the training data with two datasets: (1) 10k hours of the GigaSpeech data \cite{GigaSpeech2021}. It is collected from audiobooks, podcasts and YouTube, covering both read and spontaneous speaking styles, and a variety of topics, such as arts, science, sports, etc. 
It should be noted that the total data size of GigaSpeech is 40k, but 30k of them are not well processed. For example, there is a large segment of silence at the beginning or at the end of some utterances in the 30k data. More seriously, some utterances just contain background noise without any speech. 
Thus, we just use the subset of 10k hours of GigaSpeech data, which is well processed and validated with a segmentation pipeline proposed in \cite{GigaSpeech2021}.
(2) VoxPopuli data \cite{wang2021voxpopuli}). It is a large-scale multi-lingual unlabeled audio dataset consisting of over 400k hours of audio in 23 languages,  which is collected from 2009-2020 European Parliament (EP) event recordings including plenary sessions, committee meetings, and other events. Since our focus is English-only audio, we use 24k hours of English data in VoxPopuli for pre-training. 
In total, we collect 94k hours of data, including LibriLight, VoxPopuli, and GigaSpeech.  We believe the enriched dataset can improve the model robustness as it contains diverse audio backgrounds, more speakers, and different contents.  We call the dataset Mix 94k hr to make the description simple.

\subsection{Stabilization of Training}
Currently, it is a common practice to use 16-bit float precision (fp16) or mixed precision to pre-train large models for faster computation and less GPU memory consumption. Unfortunately, the training is unstable for large models due to the overflow issue 
(characterized by NaN losses) \cite{ding2021cogview}. A major reason is the attention score $\frac{ \mathbf{q}_i \cdot \mathbf{k}_j }{\sqrt{d}}$ is larger than the upper bound value of the fp16, resulting in the overflow issue in training. 

We apply a simple trick to alleviate the overflow issue \cite{ding2021cogview}. Given that softmax is invariant under translation by the same value in each coordinate i.e. $ \text{softmax}(\mathbf{x}+ \alpha)_k = \text{softmax}(\mathbf{x} )_k$,
where $\alpha$ denotes a constant number, the equation (\ref{eq3}) can be implemented as 
\begin{equation}
\begin{split}
    \alpha_{i,j} & \propto \exp\{\frac{ \mathbf{q}_i \cdot \mathbf{k}_j }{\sqrt{d}} + r_{i-j} \} \\
    & = \exp \{(\frac{ \mathbf{q}_i}{c \sqrt{d}} \cdot \mathbf{k}_j - \max_{j' \leq T}(\frac{ \mathbf{q}_i}{c \sqrt{d}} \cdot \mathbf{k}_{j'})) \times c + r_{i-j} \}. 
\end{split}
\end{equation}
where $c$ is a scale hyperparameter and set to 32 in our work. In this way, the overflow issue could be solved, since $ \max_{j' \leq T}(\frac{ \mathbf{q}_i}{c \sqrt{d}} \cdot \mathbf{k}_{j'})$ could guarantee the max value is smaller than $2^{16}$. 

\section{Experiment}
\subsection{Pre-Training Setup} \label{setup}
The WavLM Base and WavLM Base+ have 12 Transformer encoder layers, 768-dimensional hidden states, and 8 attention heads, resulting in 94.70M parameters. The WavLM Large has 24 Transformer encoder layers, 1024-dimensional hidden states, and 12 attention heads, resulting in 316.62M parameters. The relative position embeddings are shared across all layers, which avoids significantly increasing the number of parameters. 
We pre-train the WavLM Base model for 400k steps on LibriSpeech 960 hours audio \cite{librispeech} using the label generated by clustering the 6-th transformer layer output of the 1st-iteration HuBERT Base model.
The WavLM Base+ and WavLM Large models are pre-trained for 1M and 700k steps on 94k large-scale diverse data using the labels generated by clustering the 9th transformer layer output of the released 2nd-iteration HuBERT Base model \footnote{\url{https://github.com/pytorch/fairseq/tree/main/examples/hubert}}. Even if the 1st iteration WavLM is better than HuBERT, we find using the 1st iteration pseudo-label of HuBERT results in a slightly better 2nd iteration WavLM (0.2 WER reduction on LibriSpeech dev-other). 
The masked speech denoising  modeling is considered in 20$\%$ utterances, where the mixing noise probability $p_n$ is set to 0 for WavLM Base model, and 10$\%$ for WavLM Base+ and Large model.
For other training configurations, the same hyperparameters are used following \cite{hsu2021hubert}, which are shown in Appendix~\ref{ssec:hyper_pretrain}. 

\begin{table*}[ht]
\centering
\caption{
Universal speech representation evaluation on SUPERB benchmark. ParaL denote Paralinguistics aspect of speech.  \label{table:exp}
}
\resizebox{1.0\textwidth}{!}{
\begin{tabular}{l|c|c||r|r|r||r|r|r|r|r||r|r|rr||r||rr|r|rrr||c}

\hline
\multirow{3}{*}{Method} & \multirow{3}{*}{\#Params} & \multirow{3}{*}{Corpus}
 & \multicolumn{3}{c||}{Speaker} & \multicolumn{5}{c||}{Content} & \multicolumn{4}{c||}{Semantics}  & ParaL & \multicolumn{6}{c||}{Generation} & Overall \\ \cline{4-23}

& & & SID & ASV & SD & PR & ASR & OOD-ASR & KS & QbE & ST & IC& \multicolumn{2}{c||}{SF} & ER & \multicolumn{2}{c|}{SE} & SS & \multicolumn{3}{c||}{VC} \\ \cline{4-23}

& & & Acc $\uparrow$ & EER $\downarrow$ & DER $\downarrow$ & PER $\downarrow$ & WER $\downarrow$ & WER $\downarrow$ & Acc $\uparrow$ & MTWV $\uparrow$  & BLEU $\uparrow$ & Acc $\uparrow$ & F1 $\uparrow$ & CER $\downarrow$ & Acc $\uparrow$ & PESQ $\uparrow$ & STOI $\uparrow$ & SI-SDRi $\uparrow$ & MCD $\downarrow$ & WER $\downarrow$ & ASV $\uparrow$ & Score $\uparrow$ \\ \hline 

FBANK & 0 & - & 8.5E-4 & 9.56 & 10.05 & 82.01 & 23.18 & 63.58 & 8.63 & 0.0058 &  2.32 & 9.10 & 69.64 & 52.94 & 35.39  &  2.55 & 93.6 & 9.23 & 8.47 &  38.3 & 77.25 & 43.2 \\ \hline

PASE+~\cite{pase+} & 7.83M & LS 50 hr & 37.99 & 11.61 & 8.68 & 58.87 & 25.11 & 61.56 & 82.54 & 0.0072 &  3.16 & 29.82 & 62.14 & 60.17 & 57.86 & 2.56	 &93.9	 &9.87	 & 8.66	 &30.6	 &	63.20 &  51.5 \\ \hline

APC~\cite{apc1} & 4.11M & LS 360 hr & 60.42 & 8.56 & 10.53 & 41.98 & 21.28 & 63.12  & 91.01 & 0.0310 & 5.95 & 74.69 & 70.46 & 50.89 & 59.33 & 2.56	 & 93.4	 &  8.92	 & 8.05	 & 27.2	 & 87.25	 &  59.2\\

VQ-APC~\cite{vq_apc} & 4.63M & LS 360 hr & 60.15 & 8.72 & 10.45 & 41.08 & 21.20 & 63.56 & 91.11 & 0.0251 & 4.23 & 74.48 & 68.53 & 52.91 & 59.66 & 2.56	 &	93.4  &	  8.44 &7.84	 &22.4	 &	94.25 &  59.5\\

NPC~\cite{npc} & 19.38M & LS 360 hr & 55.92 & 9.40 & 9.34 & 43.81 & 20.20 & 61.66  & 88.96 & 0.0246 &  4.32 & 69.44 & 72.79 & 48.44 & 59.08 &  2.52	 & 93.1	 &	 8.04 &	7.86 & 30.4	 & 94.75	 &  59.0 \\

Mockingjay~\cite{mockingjay} & 85.12M & LS 360 hr & 32.29 & 11.66 & 10.54 & 70.19 & 22.82  & 65.27 & 83.67 & 6.6E-04 & 4.45 & 34.33 & 61.59 & 58.89 & 50.28 & 2.53	 &	93.4 &	 9.29 &  8.29	 &  35.1	 &	79.75 & 51.0  \\

TERA~\cite{tera} & 21.33M & LS 960 hr & 57.57 & 15.89 & 9.96 & 49.17 & 18.17  & 58.49  & 89.48 & 0.0013 & 5.66  & 58.42 & 67.50 & 54.17 & 56.27 &	2.54 &93.6	 & 10.19	 & 8.21	 & 25.1	 &	83.75 &  57.2 \\

DeCoAR 2.0~\cite{decoar2} & 89.84M & LS 960 hr & 74.42 & 7.16 & 6.59 & 14.93 & 13.02 & 53.62 & 94.48 & 0.0406 & 9.94 & 90.80 & 83.28 & 34.73 & 62.47  & 2.47	 & 93.2 &	 8.54& 7.83	 & 17.1	 & 	90.75 & 66.3 \\

\hline

modified CPC~\cite{modified_cpc} & 1.84M & LL 60k hr & 39.63 & 12.86 & 10.38 & 42.54 & 20.18 & 62.54  & 91.88 & 0.0326 &  4.82 & 64.09 & 71.19 & 49.91 & 60.96 &2.57	 &93.7	 &10.40	 &8.41	 &26.2	 &	71.00 & 56.9\\

wav2vec~\cite{wav2vec} & 32.54M & LS 960 hr & 56.56 & 7.99 & 9.9 & 31.58 & 15.86 & 55.86  & 95.59 & 0.0485 &  6.61 & 84.92 & 76.37 & 43.71 & 59.79 &  2.53	 &93.8	 &	 9.30 &	7.45 &	10.1 & 98.25	 & 63.5 \\

vq-wav2vec~\cite{vq_wav2vec} & 34.15M & LS 960 hr & 38.80 & 10.38 & 9.93 & 33.48 & 17.71 & 60.66 & 93.38 & 0.0410 & 5.66 & 85.68 & 77.68 & 41.54 & 58.24 &2.48	 &	 93.6 &	8.16 &\textbf{7.08}	 &13.4	&\textbf{100.00}	 & 61.8 \\

wav2vec 2.0 Base~\cite{wav2vec2} & 95.04M & LS 960 hr & 75.18  & 6.02 & 6.08 & 5.74 & 6.43 & 46.95 & 96.23 & 0.0233 & 14.81 & 92.35 & 88.30 & 24.77 & 63.43 	 &2.55	 &	  93.9 &	9.77 &7.50	 &10.5	&98.00	 & 69.6\\

HuBERT Base~\cite{hsu2021hubert} & 94.68M & LS 960 hr & 81.42 & 5.11 & 5.88  & 5.41 & 6.42 &46.69 & 96.30 & 0.0736 & 15.53 & 98.34 & 88.53 & 25.20 & 64.92 &	 2.58 &	93.9 &9.36	 &	 7.47 &	 8.0&98.50	 & 70.9\\

WavLM Base &   94.70M & LS 960 hr &84.51&	4.69&	4.55&	4.84&	6.21	&	42.81 	&	96.79&	0.0870& 20.74 &	98.63&	89.38&	22.86 & 65.94&	2.58 &	94.0 &	10.37 &	7.42		 &8	 &98.00	 & 72.0
 \\
~~- w/o denoising task &  94.70M & LS 960 hr& 84.39&	4.91&	6.03&	4.85&	6.08&43.61	 &	96.79&	0.0799& 21.03 &	98.42&	88.69	&23.43	&65.55&	 2.56 & 93.9	 &	9.91 &	7.43 &\textbf{7.5}	 &97.75	 & 71.7\\
~~- w/o structure modification &  94.68M & LS 960 hr& 84.74&	4.61&	4.72&	5.22&	6.80& 42.88	 &	96.79&	0.0956& 20.03 &	98.31&	88.56	&24.00	&65.60&	2.58 &	94.0 & 10.29	&	7.45 &	8.4 &99.00	  & 71.9 \\

WavLM Base+ & 94.70M & Mix 94k hr &89.42&	4.07&	3.50&	3.92&	5.59& 38.32	 &	97.37&	\textbf{0.0988}	& 24.25 & 99.00 & 90.58	& 21.20 &	68.65 &	2.63 &94.3	 &	10.85 &	 7.40 &	8.1 &	99.00 & 73.4

\\
\hline

wav2vec 2.0 Large~\cite{wav2vec2} & 317.38M & LL 60k hr& 86.14  & 5.65 &  5.62 & 4.75 & 3.75 & 44.69 & 96.66 & 0.0489 &12.48	 & 95.28 & 87.11 & 27.31 & 65.64  &2.52	 &	 94.0 &10.02	 &7.63	 &	15.8 &	 97.25 & 70.4 \\

HuBERT Large~\cite{hsu2021hubert} & 316.61M & LL 60k hr & 90.33 & 5.98 & 5.75 & 3.53 & 3.62 & 44.08  & 95.29 & 0.0353 & 20.01 &98.76 & 89.81 & 21.76 & 67.62  &2.64	 &94.2	 &10.45	 &7.22	 & 9.0	 & 99.25 & 72.2 \\

WavLM Large & 316.62M & Mix 94k hr & \textbf{95.49}	& \textbf{3.77} &	\textbf{3.24}	& \textbf{3.06}& \textbf{3.44}	& \textbf{32.27}	&	\textbf{97.86}&	0.0886& \textbf{26.57} &	\textbf{99.31} &	\textbf{92.21}& \textbf{18.36}&	 \textbf{70.62} &\textbf{2.70}	 &\textbf{94.5}	 &\textbf{11.19}	 &	7.30 &9.0	 &99.00	 & \textbf{74.6}
 \\

\hline
\end{tabular}}

\end{table*}

\raggedbottom

\subsection{Universal Representation Evaluation}

\subsubsection{Setup}
We first evaluate our models on SUPERB, which is designed to provide 
a standard and comprehensive testbed for pre-trained models on various speech tasks. It covers fifteen tasks, including Speaker Identification (SID),	 Automatic Speaker Verification (ASV), Speaker Diarization (SD), Phoneme Recognition (PR), Automatic Speech Recognition (ASR),  Out-Of-Domain Automatic Speech Recognition (OOD-ASR), Keyword Spotting (KS), Query by Example Spoken Term Detection (QbE),  Speech Translation (ST), Intent Classification (IC), Slot Filling (SF), Emotion Recognition (ER), Speech Enhancement (SE), Speech Separation (SS) and Voice Conversion (VC). These tasks can be grouped into five aspects
of speech: content, speaker, semantics, paralinguistics, and generation.

We follow the settings created by SUPERB. 
1) We use the same downstream models as the SUPERB implementations for each downstream task.
2) Pre-trained models are frozen to limit the space of the fine-tuning hyperparameter search. 
3) The downstream models consume the weighted sum results of the hidden states extracted from each layer of the pre-trained model. 
The detailed hyperparameters for fine-tuning our WavLM models on SUPERB downstream tasks are shown in Appendix~\ref{ssec:hyper_downstream}.
The overall score is computed by ourselves: we multiply the QbE score with 100, replace each error rate score with (1 - error rate), and average the scores of all tasks.

\subsubsection{Evaluation result}
Table \ref{table:exp} shows the evaluation results. We compare our WavLM with several SSL models which are evaluated by \cite{superb,tsai2022superb}. In general, WavLM is very powerful in universal representation learning. Our WavLM Base+ model has outperformed HuBERT large and wav2vec 2.0 large in the overall score. 

\textbf{WavLM Base}:
From Table \ref{table:exp}, we can observe that WavLM Base performs better than wav2vec 2.0 Base and HuBERT Base on all downstream tasks. It is a fair comparison as the three models use the same amount of pre-training data and the number of parameters. The results indicate the effectiveness of our structure and the masked speech denoising  modeling  in universal speech representation learning. 
We find the most impressive result is speaker diarization, where the WavLM Base outperforms HuBERT Base by 22.6\% relatively. Our explanation is that the additional overlapped speech forces the model to deal with multi-speaker signals during pre-training. 
To verify this assumption, we conduct an ablation study to remove simulated noisy/overlapped speech in pre-training. The performance of the ``w/o denoising task" drops significantly for the speaker diarization task. We also evaluate the contribution of the structure change. We can see that, in the ``w/o structure modification" setting, performance degradation can be witnessed especially for PR and ASR tasks. It indicates that the gated relative position bias contributes to the performance improvement of the content-related tasks. Meanwhile, we can observe that WavLM performs very well on semantic, paralinguistics, and generation tasks as well, demonstrating our model is general for the full stack speech processing tasks. 

\textbf{WavLM Base+:} WavLM Base+ shows the contribution from larger and more diverse pre-training data. It consistently improves WavLM Base and even outperforms the wav2vec 2.0 Large and HuBERT Large in the overall score. This indicates that the 960h data are insufficient to fulfill the capacity of the Base model. The combined dataset especially boosts the performance of the testsets which are not extracted from the audiobook, such as ASV, OOD-ASR, IC, SF, and ER.

\textbf{WavLM Large}:  Most tasks benefit from the larger model size, especially for the ASR. We obtain 38$\%$ word error rate reduction on the ASR by model scaling-up. Furthermore, there is 6.07$\%$ absolute improvement on the SID task, indicating the large model size also impacts the speaker-related tasks. Compared to the HuBERT Large model, WavLM Large is consistently better across 14 downstream tasks, demonstrating the modifications are effective for the large-scale models. 

\begin{figure*}[tbp]
	\centering
	\subfloat[\footnotesize HuBERT Base]
	{\label{sfig:hubert_base}\includegraphics[width=0.45\textwidth,trim=25 50 68 90,clip]{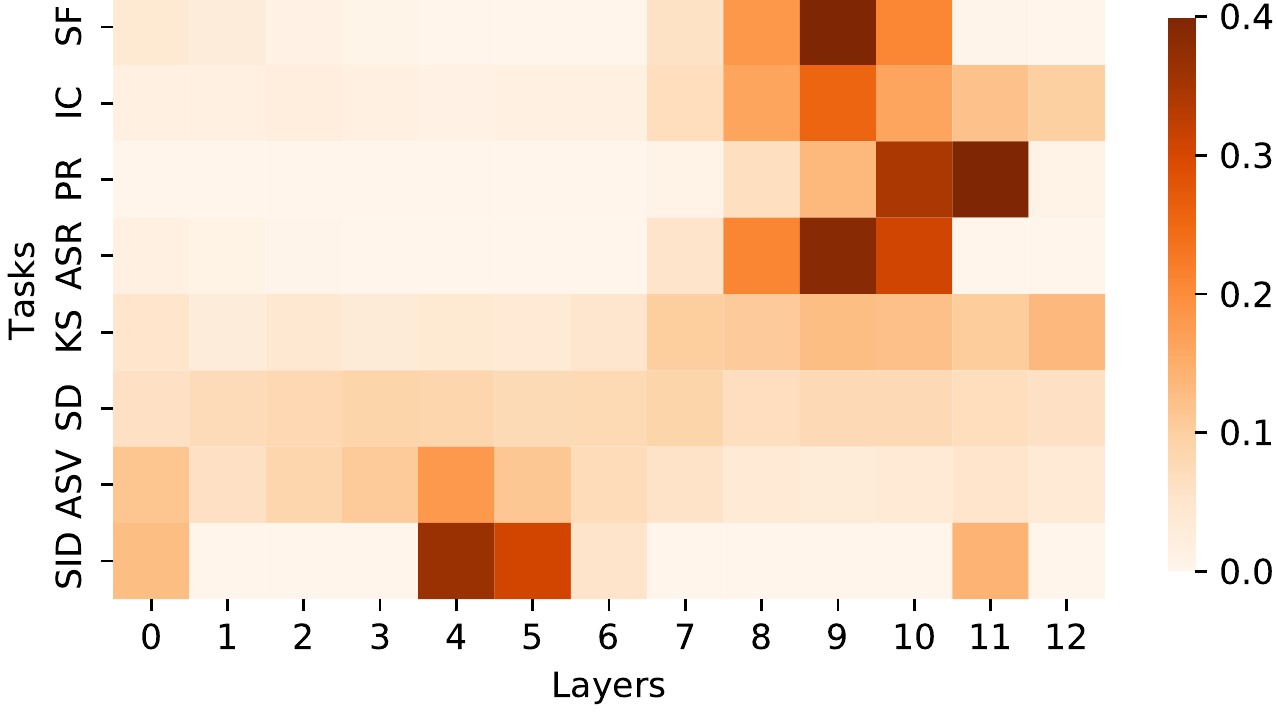}}\quad
	\subfloat[\footnotesize WavLM Base+] 
	{\label{sfig:wavlm_base_plus}\includegraphics[width=0.45\textwidth,trim=25 50 68 90,clip]{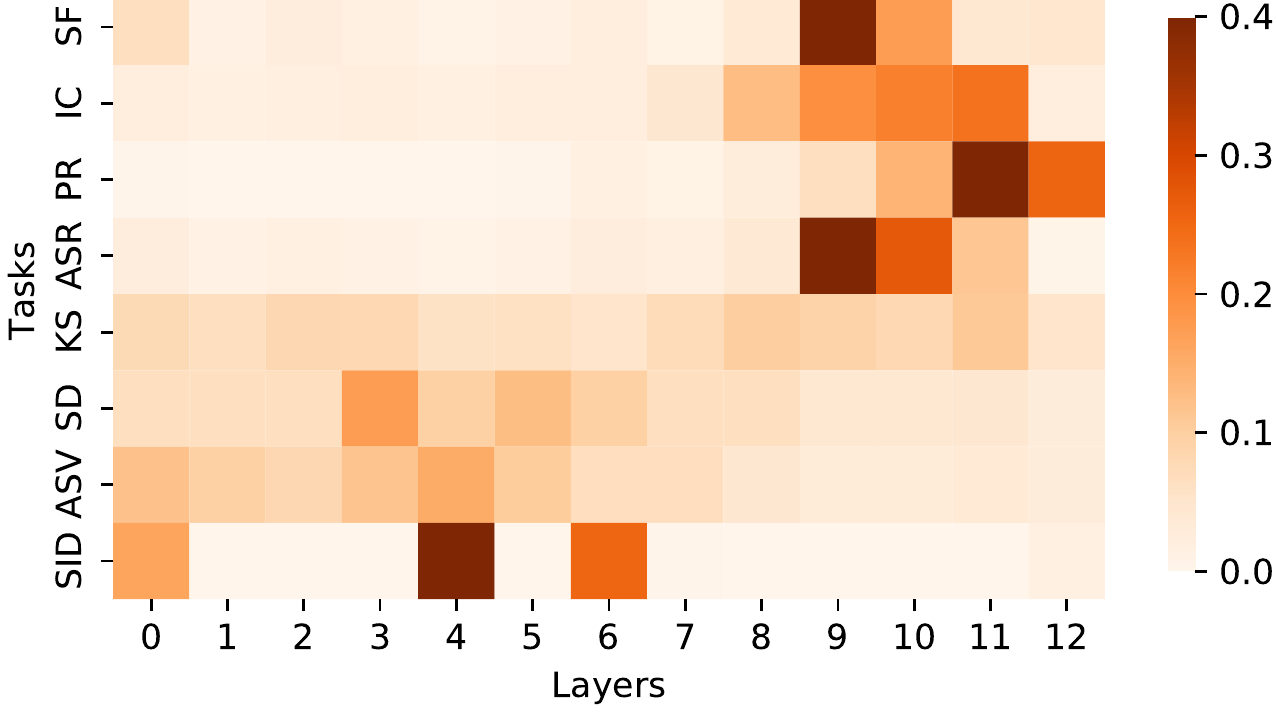}}\\
	\subfloat[\footnotesize HuBERT Large]	
	{\label{sfig:hubert_large}\includegraphics[width=0.45\textwidth,trim=100 160 200 200,clip]{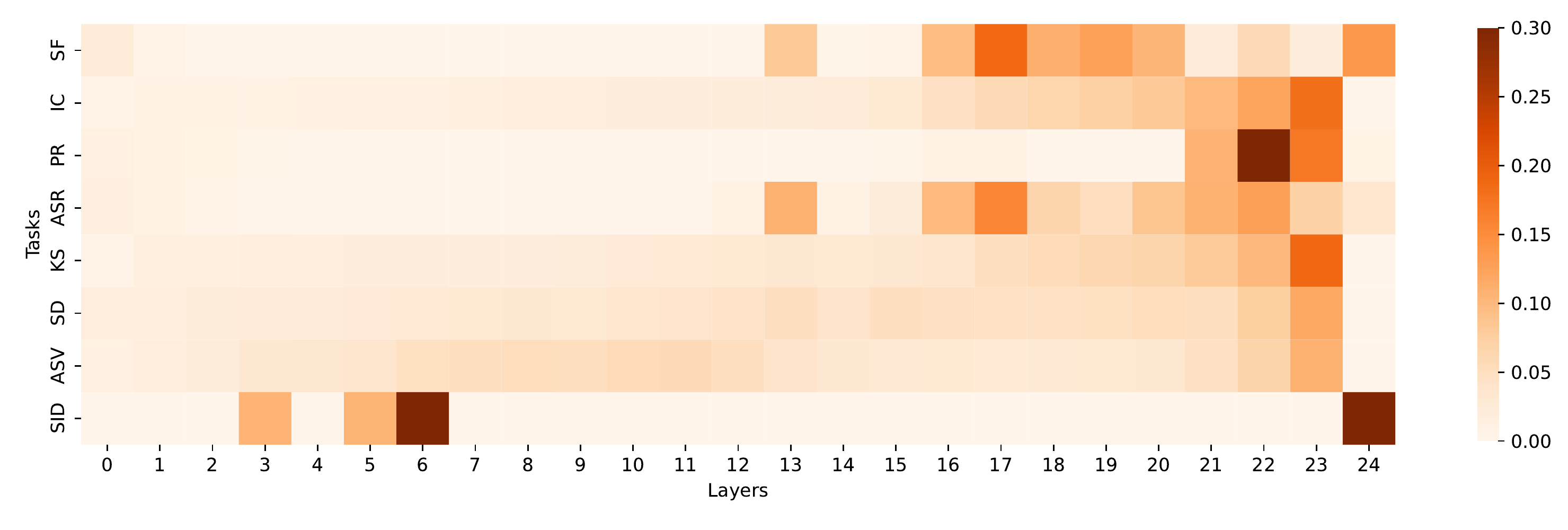}}\quad
	\subfloat[\footnotesize WavLM Large] 
	{\label{sfig:wavlm_large}\includegraphics[width=0.45\textwidth,trim=100 160 200 200,clip]{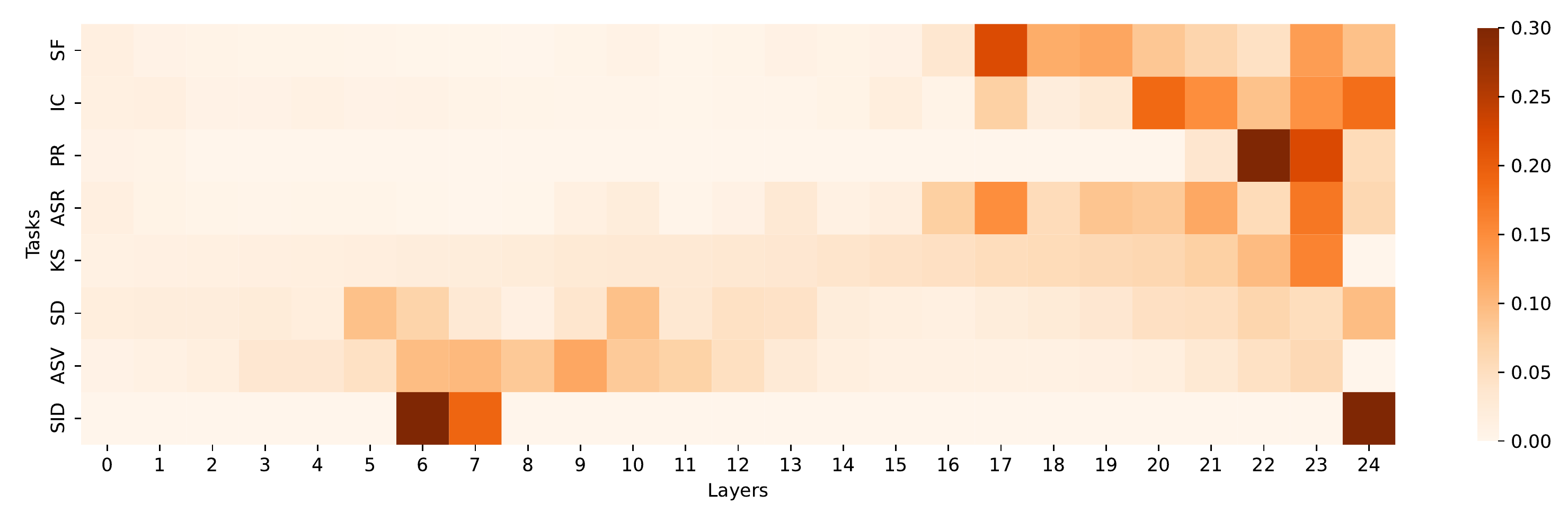}}\\	
	\caption{Weight analysis on the SUPERB Benchmark. Layer 0 corresponds to the input of the first Transformer layer. The y-axis represents different tasks, while the x-axis represents different layers. 
	}\label{fig:weight-anaysis}
\end{figure*}

\subsubsection{Analysis}
\label{superb:analysis}

Following the SUPERB policies, we weighted-sum the hidden states of different layers and feed it to the task-specific layers. 
Figure~\ref{fig:weight-anaysis} shows the weights of different layers of HuBERT and WavLM models on the different downstream tasks of the SUPERB benchmark. The larger weight indicates the greater contribution of the corresponding layer.  We normalize the weights from different layers based on the hidden state values of their corresponding layers, which eliminates the weight bias to layers with smaller hidden state values.  

As for the Base models, the contribution patterns of different layers are similar between WavLM and HuBERT, as shown in Figure \ref{sfig:hubert_base} and \ref{sfig:wavlm_base_plus}. We can observe that the bottom layers contribute more to speaker-related tasks, such as speaker identification, automatic speaker verification, and speaker diarization. On the other hand, for automatic speech recognition, phoneme recognition, intent classification, and slot filling tasks, the top layers are more important. It indicates the Base models learn speaker information with the bottom layers while the content and semantic information are encoded in the top layers. The model behavior is similar to Large models. In  Figures \ref{sfig:hubert_large} and \ref{sfig:wavlm_large}, we can see that the top layers contribute most to content and semantic tasks, while the middle layers have a great impact on speaker tasks. The phenomenon indicates how to leverage hidden states of middle layers is the key to the success of speaker-related tasks.  

Since SUPERB requires the pre-trained model frozen in fine-tuning, it cannot show the power of pre-trained models. To explore the limit of our models, we further select typical speech tasks to evaluate our pre-trained model performance. Four tasks are used to evaluate our model from different perspectives, and the training data amount is not on the same scale for the four tasks. 
The details of the tasks can be found in Appendix \ref{ssec:set_downstream}.

\subsection{Speaker Verification }
\label{ssec:sv}

\subsubsection{problem formulation}
The training dataset for speaker verification contains audio and speaker id pairs as $\mathcal{D} = \{\mathbf{x_i},\mathbf{y_i}\}$. Given audio clip $\mathbf{x}$ and a reference $\mathbf{x}'$, the goal of speaker verification is to determine whether $\mathbf{x}'$ is from the same speaker as $\mathbf{x}$. 

\subsubsection{Datasets}
VoxCeleb1 \cite{nagrani2017voxceleb} and VoxCeleb2 \cite{chung2018voxceleb2} datasets are used in our experiments for speaker verification.   For data pre-processing, we apply online data augmentation using the MUSAN \cite{snyder2015musan} noise, DNS noise \cite{reddy2021interspeech} and the RIR \footnote{\url{https://www.openslr.org/28/}} reverberation with probability 0.6. Voice activity detection (VAD) processing is not adopted. We use all three official trial lists Vox1-O, Vox1-E, and Vox1-H to evaluate the system. 

\subsubsection{Setup}
\label{sssec:asv_implementation}
We choose the ECAPA-TDNN (small) \cite{desplanques2020ecapa} architecture as the downstream model and compare different input speech representations, including handcrafted features and the pre-training features. The model contains a frame encoder to extract speaker information from the input sequence, a statistic pooling layer to transform input to a fixed-dimensional representation, and a fully connected layer to extract speaker embedding.  For the handcrafted feature, we compare the reported results in \cite{desplanques2020ecapa} with our re-implemented results, where we extract the 40-dimensional Fbank feature with 25ms window size and 10ms frameshift. For pre-trained representations, we compare  WavLM with HuBERT model. Following SUPERB evaluation, we weighted-sum the representations from different transformer layers with learnable weights as the input to the downstream speaker verification task.  
 
In the training stage, all the recordings are chunked into 3s segments to construct the training batches. We use the additive angular margin (AAM) loss \cite{deng2019arcface} for model optimization and set the margin to 0.2. We also add an Inter-TopK penalty \cite{zhao2021speakin} on the 5 easily misclassified centers with a penalty margin of 0.1. We train the ECAPA-TDNN system with Fbank feature for 165 epochs. For systems using pre-trained representations, we first fix the pre-trained model to train ECAPA-TDNN for 20 epochs and then finetune both the pre-trained and ECAPA-TDNN  models for another 5 epochs. When we add the large margin fine-tuning strategy \cite{thienpondt2021idlab}, we train the systems for an extra 2 epochs, during which we sample 6s training segments and set the AAM margin to 0.4.

In the evaluation stage, the whole utterance is fed into the system to extract speaker embedding. We use cosine similarity to score the evaluation trial list. We also use the adaptive s-norm \cite{karam2011towards,cumani2011comparison} to normalize the trial scores. The imposter cohort is estimated from the VoxCeleb2 dev set by speaker-wise averaging all the extracted speaker embeddings. We set the imposter cohort size to 600 in our experiment. To further push the performance, we also introduce the quality-aware score calibration \cite{thienpondt2021idlab} for our best systems, where we randomly generate 30k trials based on the VoxCeleb2 test set to train the calibration model.

\begin{table}[]
    \centering
    \caption{Speaker verification results on VoxCeleb1. For the lines with $^*$ notation, we add the large margin fine-tuning and quality-aware score calibration \cite{thienpondt2021idlab} to push the limit of the performance. 
    }
    \label{tab:asv_result}
    \resizebox{\columnwidth}{!}{
    \begin{tabular}{l|ccc} 
         \toprule 
         \multirow{2}{*}{Feature} & \multicolumn{3}{c}{EER (\%)} \\
         & Vox1-O & Vox1-E & Vox1-H \\
         \hline
         ECAPA-TDNN \cite{desplanques2020ecapa} & 1.010 & 1.240 & 2.320 \\
         ECAPA-TDNN  (Ours) & 1.080 & 1.200 & 2.127 \\
         \hline
         HuBERT Base & 0.989 & 1.068 & 2.216\\
         HuBERT Large & 0.808 & 0.822 & 1.678\\
         WavLM Base+ & 0.84 & 0.928 & 1.758\\
         WavLM Large & 0.617 & 0.662 & 1.318\\
         \hline
          HuBERT Large$^*$ & 0.585 & 0.654 & 1.342\\
          WavLM Large$^*$ & \textbf{0.383} & \textbf{0.480} & \textbf{0.986} \\
         \bottomrule
       
    \end{tabular}
    }
\end{table}

\begin{table*}[!th]
\centering
\caption{Diarization error rate (DER \%) results on CALLHOME with estimated number of speakers.}
\begin{adjustbox}{width=.8\textwidth,center}
\begin{threeparttable}
\begin{tabular}{l|ccccc|c} 
         \toprule 
         \multirow{2}{*}{Method} & \multicolumn{6}{c}{\# of speakers in a session} \\
          & 2 & 3 & 4 & 5 & 6 & all\\
          \hline
          x-vector clustering \cite{horiguchi2020end} &  15.45 & 18.01 & 22.68 & 31.40 & 34.27 & 19.43 \\
          SC-EEND \cite{fujita2020neural} $^\ddagger$ & 9.57 & 14.00 & 21.14 & 31.07 & 37.06 & 15.75 \\
          VBx \cite{landini2022bayesian} $^\dagger$ $^\ddagger$ & 9.44 & 13.89 & 16.05 & 13.87 & 24.73 & 13.28 \\
          EEND-EDA \cite{horiguchi2021encoder} & 8.50 & 13.24 & 21.46 & 33.16 & 40.29 & 15.29 \\
          EEND-EDA clustering \cite{horiguchi2021towards} & 7.11 & 11.88 & 14.37 & 25.95 & 21.95 & 11.84 \\
          EEND-vector clustering \cite{kinoshita2021advances} & 7.96 & 11.93 & 16.38 & 21.21 & 23.10 & 12.49 \\
          EEND-vector clustering (Ours) & 7.54 & 12.42 & 18.41 & 26.79 & 27.40 & 13.31 \\
          \hline
          HuBERT Base \& EEND-vector clustering & 7.93 & 12.07 & 15.21 & 19.59 & 23.32 & 12.63 \\
              HuBERT Large \& EEND-vector clustering & 7.39 & 11.97 & 15.76 & 19.82 & 22.10 & 12.40 \\
          WavLM Base+  \& EEND-vector clustering& 6.99 & 11.12 & 15.20 & 21.61 & 21.70 & 11.78 \\
          WavLM Large  \& EEND-vector clustering& \textbf{6.46} & \textbf{10.69} & \textbf{11.84} & \textbf{12.89} & \textbf{20.70} & \textbf{10.35} \\

         \bottomrule
    \end{tabular}
\begin{tablenotes}\footnotesize
\item $^\dagger$ Oracle speech segments were used.
\item $^\ddagger$ Results for these systems are provided in \cite{kinoshita2021advances}.
\end{tablenotes}
\end{threeparttable}
\label{tab:diarization_result}
\end{adjustbox}
\end{table*}
\raggedbottom

\subsubsection{Results}
 Table \ref{tab:asv_result} shows the results for the speaker verification task. 
From the results, we find that all the systems with pre-trained representations exceed the Fbank baseline system on the Vox1-O and Vox1-E trials. The system with HuBERT Base representations is slightly worse than the Fbank feature on the Vox1-H trial. Interestingly, the representations extracted from our proposed pre-trained models, WavLM Base+ and Large, both outperform the SOTA ECAPA-TDNN system. Compared with the Fbank feature, the representations from WavLM Large achieve over 35\% relative EER improvement on all three trials for the VoxCeleb1 evaluation set. To further push the limit of the speaker verification system, we introduce the large margin fine-tuning and quality-aware score calibration strategies \cite{thienpondt2021idlab} into our best systems and the corresponding results are listed at the bottom of Table \ref{tab:asv_result}. With these two strategies, our best system exceeds the winner system \cite{zhao2021speakin} (Vox1-O: 0.461, Vox1-E: 0.634, Vox1-H: 0.993) in VoxSRC challenge 2021\footnote{\url{https://www.robots.ox.ac.uk/~vgg/data/voxceleb/interspeech2021.html}} on all the three trials.

\subsection{Speaker Diarization }
\label{ssec:sd}

\subsubsection{problem formulation}
Speaker diarization is the task to answer ``Who spoke when?". Given a speech recording $ \mathbf{x} = (x_1, ..., x_T)$, we should assign one or more labels to each $x_t$ according to the speaker identity. When we assign more than one label to $x_t$, it indicates more than one person is speaking at time $t$, i.e. speaker overlap. Normally, we cannot know the number of speakers of a whole recording in advance. Thus, the built diarization system should have the ability to predict the speaker number for the whole recording and the speaker labels for each frame at the same time.

\subsubsection{Datasets}
The dataset used in our experiments is split into two parts. The first part is the large-scale simulation training data. The second part is the real data, which is used for evaluation and adaptation. Following the data simulation setup in \cite{kinoshita2021advances}, all the speech data from Switchboard-2 (Phase I \& II \& III), Switchboard Cellular (Part 1 \& 2), the NIST Speaker Recognition Evaluation (2004 \& 2005 \& 2006 \& 2008), the noises from  \cite{snyder2015musan} and the simulated room impulse responses used in \cite{ko2017study} are leveraged for multi-talker speech simulation. Based on the simulation pipeline introduced in \cite{fujita2019end}, we generate almost 7000 hours simulation data by setting $N_{spk}=3$ and $\beta = 10$.
We use the telephone conversation dataset CALLHOME \cite{callhome} for evaluation and adaptation. CALLHOME dataset has 500 sessions of multilingual telephonic speech where each session contains 2 to 6 speakers. Following the data usage in \cite{kinoshita2021advances}, we split the CALLHOME dataset into two parts. The first part is used for adaptation and the second part is used for evaluation.

\subsubsection{Implementation Details}
We leverage the system in \cite{kinoshita2021advances} as our downstream speaker diarization model. In the system, a long-form recording is first segmented into short blocks, where each short block is assumed to contain at most $S_{Local}$ speakers. As with \cite{kinoshita2021advances}, we set $S_{Local}=3$ in our experiment. Then, the Mel-filterbank-based features extracted from each short block are fed into a Transformer encoder to get the diarization results and $S_{Local}$ speaker embeddings. With the predicted diarization results and estimated speaker embeddings, the whole system is trained by a diarization loss and a speaker loss. During the inference, the diarization results and speaker embeddings are first predicted for each block. A clustering method is then applied to associate the embeddings from the same speaker but in different blocks.

Following the implementation in \cite{kinoshita2021advances}, we set the block length to 15s, 30s, 30s for the training, adaptation, and evaluation stage respectively. The constrained AHC (Agglomerative Hierarchical Clustering) method is used for embedding clustering during the evaluation stage. When leveraging the pre-trained representations, as with the implementation in section \ref{sssec:asv_implementation}, we just replace the handcrafted Fbank feature with the pre-trained representation $\mathbf{H}$.
Unlike \cite{kinoshita2021advances}, when we feed the diarization system with the pre-trained representations, we do not concatenate the context features for each frame and do not apply the 10 times down-sampling. We find that updating the parameters of the pre-trained model does not improve performance on the CALLHOME dataset. Thus, we freeze the pre-trained model in the fine-tuning stage.  One possible explanation is that the test data are real recordings while the training data are simulated recordings, and the model is over-fitted if the pre-trained model is not frozen.

\subsubsection{Results}
The speaker diarization results on CALLHOME dataset are shown in Table \ref{tab:diarization_result}. In our experiment, we try to reproduce the system in \cite{kinoshita2021advances} but get slightly worse results. When we replace the handcrafted feature with pre-trained representations, all the systems exceed the performance of our implemented EEND-vector clustering. Compared with the HuBERT, the representations extracted from our proposed WavLM are more useful in speaker diarization. Our proposed WavLM Base+ even outperforms the HuBERT large model. This is because the WavLM models have seen the multi-talker and speaker-overlapped speeches during the training process, and the corresponding training strategy is designed to help WavLM better process this kind of input. Finally, we also list the CALLHOME results from some recently published works. Compared with these results, it is worth noting that our best system has surpassed all the systems evaluated on the CALLHOME dataset and achieved a new SOTA performance.

\begin{table}[tb]
\setlength{\tabcolsep}{3pt}
    \centering
    \caption{ Separation results on LibriCSS dataset. We freeze the pre-trained parameters by default for the separation task. The results denote \%WER score evaluated with E2E Transformer based ASR model \cite{wang2019semantic}. 0S and 0L are utterances with short/long inter-utterance silence. The avg is the weighted averaged WER of different overlapped testsets. 
    }
    \label{tab:sep_result}

    \begin{tabular}{l|cccccc|c} 
         \toprule 
		\multirow{2}{*}{System} &
		\multicolumn{6}{c}{Overlap ratio in \%}  \\  & 0S & 0L & 10 & 20 & 30 & 40 & avg \\  \hline
        
		Conformer \cite{chen2020continuous2}  & 5.4 &  5.0 &  7.5  & 10.7 &  13.8  & 17.1 & 10.6 \\
		Conformer (rerun)  & 4.5 & 4.4 & 6.2 & 8.5 & 11.0 & 12.6 & 8.3\\
		HuBERT Base  & 4.7 & 4.6 & 6.1 & 7.9 & 10.6 & 12.3 & 8.1 \\
		\hline
		WavLM Base+ & 4.5 & 	4.4 & 	5.6 & 	7.5 & 	9.4 & 	10.9 & 7.4 \\
		 ~~- unfreeze pre-trained parameters & 4.5 & 	4.3 & 	5.9 & 	8.3 & 	11.1 & 	12.5 & 8.2 \\
		WavLM Large & \textbf{4.2} & 	\textbf{4.1} & 	\textbf{4.8} & 	\textbf{5.8} & 	\textbf{7.4} & 	\textbf{8.5}  & \textbf{6.0} \\
		\bottomrule
    \end{tabular}
\end{table}
\raggedbottom

\subsection{Speech Separation }
\label{ssec:ss}

\subsubsection{problem formulation}
The goal of speech separation is to  estimate individual speaker signals from their mixture, where the source  signals may be overlapped with each other entirely or partially. 
Given $S$ source signals $\{ \mathbf{x}_s = (x_1, ..., x_T) \}_{s=1}^S$, the mixed signal is formulated as $\mathbf{y}=\sum_{s=1}^S \mathbf{x}_s$.
$\mathbf{X}_s$ and $\mathbf{Y}$ denote the Short-Time Fourier Transform (STFT) of the source signal and mixed signal, respectively.
Following \cite{wang2014training} and \cite{erdogan2017deep}, instead of directly predicting the source STFTs, we firstly estimate a group of masks $\{\mathbf{M}_s\}_{s=1}^S$ with a deep learning model, and then obtain each source STFT with $\mathbf{X}_s = \mathbf{M}_s \odot  \mathbf{Y} $, where $\odot$ is an elementwise product. 

\subsubsection{Datasets}
Our training dataset for the separation task consists of 219 hours of artificially reverberated and mixed utterances that are sampled randomly from WSJ1 \cite{wsj1}. Four different mixture types described in \cite{yoshioka2018multi} are included in the training set. To generate each training mixture, we randomly pick one or two speakers from  WSJ1 
and convolve each with a room impulse response (RIR) simulated with the image method \cite{imagemethod}. The reverberated signals are then rescaled and mixed with a source energy ratio between -5 and 5 dB. In addition, we add simulated isotropic noise \cite{habets2007generating} with a 0--10 dB signal to noise ratio. The average overlap ratio of the training set is around 50\%. 

LibriCSS is used for evaluation \cite{chen2020continuous}. 
The dataset has 10 hours of seven-channel recordings of mixed and concatenated LibriSpeech test utterances. The recordings were made by playing back the mixed audio in a meeting room. 
We apply the single-channel utterance-wise evaluation schemes of LibriCSS, where the long-form recordings are segmented into individual utterances by using ground-truth time marks to evaluate the pure separation performance. 

\subsubsection{Implementation details}

For the separation task, we use the previous SOTA work \cite{chen2020continuous2} as our baseline model, which uses the Conformer-based model for separation, and consists of 16 Conformer encoder layers with 4 attention heads, 256 attention dimensions, and 1024 FFN dimensions. 
A linear projection layer and sigmoid activation function are attached to the final encoder for the mask prediction. 
Given the STFT of mixed signal $\mathbf{Y}$ as the input, the separation model estimates masks $\{\mathbf{M}_s\}_{s=1}^S$, then each source signal can be obtained as  $\{\mathbf{X}_s = \mathbf{M}_s \odot  \mathbf{Y} \}_{s=1}^S$ for each speaker.

To fine-tune our pre-trained models on the separation, we use WavLM models as feature extractors and the Conformer-base architecture as the task-specific downstream model.
We begin by extracting the pre-trained representation $\mathbf{H}$ as introduced in section \ref{sssec:asv_implementation}.
Secondly, we concatenate the pre-trained representation and STFT representation in the feature dimension.
Since the window size and hop length of STFT are typically set to 400 and 160, respectively, the STFT representation $\mathbf{Y} = \{ \mathbf{Y}_t \}_{t=1}^{T'}$ has the half stride size compared to the pre-trained representation $\mathbf{H}=  \{ \mathbf{H}_t \}_{t=1}^{T'/2}$.
To match the size of the time dimension, we duplicate the pre-trained representation with $\hat{\mathbf{H}} = \{ \hat{\mathbf{H}}_t = \mathbf{H}_{\lceil \frac{t}{2} \rceil} \}_{t=1}^{T'}$, then we can concatenate the two representations $[\mathbf{Y}_t, \hat{\mathbf{H}}_t]$ in the feature dimension for each time step.
Finally, we feed the concatenated representations to the downstream model for the mask estimation.

The separation models are trained with the AdamW optimizer \cite{loshchilov2018decoupled}, where the weight decay is set to 1e-2.
We set the learning rate to 1e-4 and use a warm-up learning schedule with a linear decay, in which the number of the warm-up steps is 10,000 and the total number of the training step is 260,000. 

We follow the previous work  \cite{chen2020continuous2, chen2021don, chen21l_fastcss,wu21f_investigation} to evaluate our model with an end-to-end Transformer  based ASR models \cite{wang2019semantic}, which achieves 2.08\% and 4.95\% word error rates (WERs) for LibriSpeech test-clean and test-other, respectively.

\subsubsection{Results}

Table~\ref{tab:sep_result} shows the single-channel utterance-wise separation results on LibriCSS dataset.
Our WavLM Base+ and Large models with the frozen pre-trained parameters achieve SOTA results on all the overlap ratio settings,  outperforming the baseline results by a large margin.

We rerun the previous SOTA work \cite{chen2020continuous2} with a modified Conformer-base architecture \cite{conformerv2} and a modified training loss \cite{wu21f_investigation}, which achieve much better baseline results.
With the pre-trained representation provided by the HuBERT Base model, the performance is comparable with the baseline results for all the overlap ratios. It is because the HuBERT model is rarely optimized with speaker-overlapped speech and lacks multi-speaker modeling during pre-training.

In contrast, our WavLM Base+ with a similar model size can successfully reduce the WER scores, especially for the large overlap ratio audios. 
We find fine-tuning the parameters of the pre-trained model yields better training accuracy but worse evaluation results than freezing the pre-trained parameters for the separation task.
An explanation is that the separation model with pre-trained parameters adaptation would be over-fitted with the artificially mixed training data, and it is evaluated with a real meeting recording dataset.
With the pre-trained representation provided by our WavLM Large model, the performance on all the overlap ratio settings can be further improved. It can achieve 32.5\% relative WER score reduction for the 40\% overlap ratio cases and 27.7\% relative WER score reduction on average.

\subsubsection{Weight Analysis}

\begin{figure}[tbp]
	\centering
	\subfloat[\footnotesize WavLM Base+] 
	{\label{sfig:wavlm_base_plus1}\includegraphics[width=.45\textwidth,trim=100 190 200 220,clip]{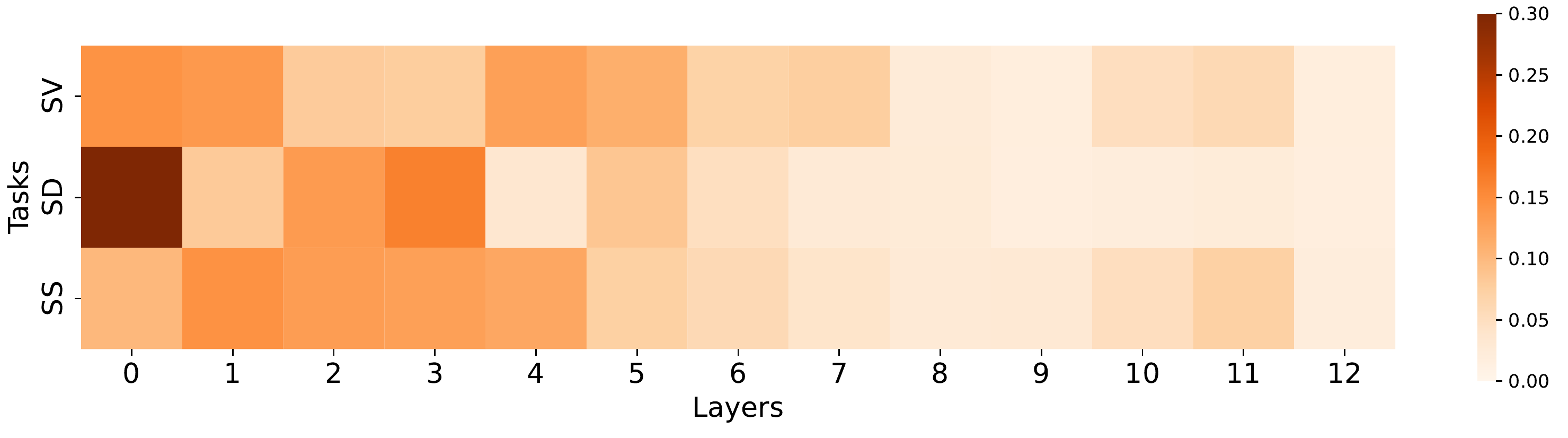}}\\
	\subfloat[\footnotesize WavLM Large] 
	{\label{sfig:wavlm_large1}\includegraphics[width=.45\textwidth,trim=100 220 220 250,clip]{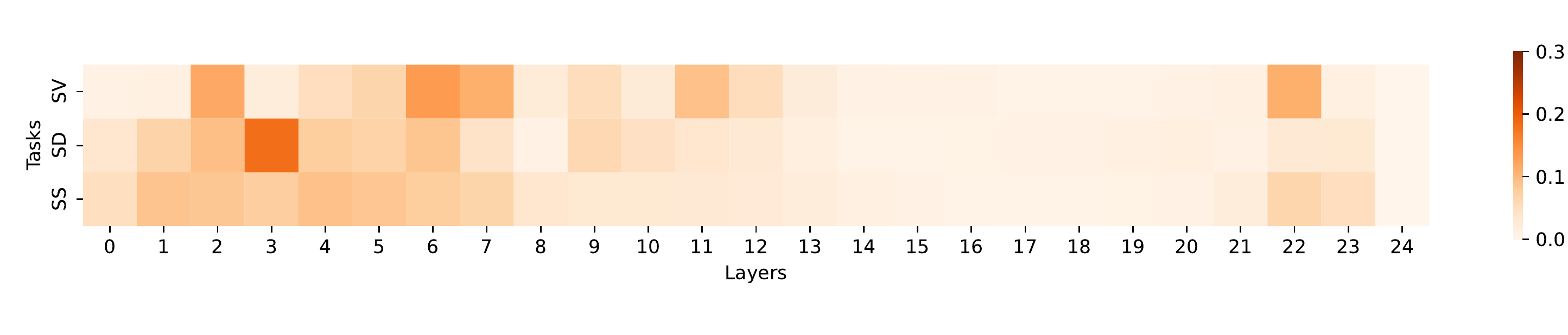}}\\	
	\caption{Weight analysis on the Speaker Verification (SV), Speech Diarization (SD) and Speech Separation (SS) tasks. Layer 0 corresponds to the input of the first Transformer layer. The y-axis represents different tasks, while the x-axis represents different layers. 
	}\label{fig:weight-anaysis-finetune-tasks}
\end{figure}

For the speaker verification (Section~\ref{ssec:sv}), speech diarization (Section~\ref{ssec:sd}) and speech separation (Section~\ref{ssec:ss}) tasks, we weighted-sum the representations from different layers of the pre-trained models as the input to the task-specific downstream models.
Figure~\ref{fig:weight-anaysis-finetune-tasks} shows the weights of different layers of WavLM Base+ and WavLM Large models on these tasks. 
As with the weight analysis on the SUPERB benchmark in Section~\ref{superb:analysis}, we can observe that the contribution mostly comes from the bottom layers for all these  tasks. It indicates that the shallow layers of WavLM models learn the speaker-related information during the SSL procedure. It is essential to leverage hidden states of intermediate layers for speaker-related tasks to make full use of the pre-trained knowledge of WavLM models. 

\begin{table}[!th]
\begin{center}
\caption{WER on LibriSpeech test sets when trained on the Libri-Light low-resource labeled data setups of 1 hour, 10 hours and the clean 100h subset of LibriSpeech.\label{asr_main_result} }
\resizebox{\columnwidth}{!}{
\begin{tabular}{lcccc}
\toprule
Model & Unlabled Data &  LM &   test-clean & test-other \\ \hline
\midrule
\multicolumn{5}{c}{\textbf{\textit{1-hour labeled}}} \\ \hline
wav2vec 2.0 Base  & LS-960 & None & 24.5 & 29.7  \\ 
WavLM Base  & LS-960 &  None & 24.5   & 29.2  \\  
WavLM Base+ & MIX-94k & None &  22.8 &  26.7  \\ \hline

DeCoAR 2.0 & LS-960 & 4-gram & 13.8 & 29.1  \\
DiscreteBERT  & LS-960 & 4-gram & 9.0 & 17.6  \\
wav2vec 2.0 Base  & LS-960 & 4-gram  & 5.5 & 11.3 \\
HuBERT Base   & LS-960 & 4-gram &  6.1 & 11.3\\
WavLM Base  & LS-960 &  4-gram &  5.7  &  10.8 \\ 
WavLM Base+ & MIX-94k & 4-gram &  5.4  &   9.8 \\ 

wav2vec 2.0 Large  & LL-60k & 4-gram  & 3.8 & 7.1 \\
WavLM Large & MIX-94k & 4-gram & 3.8 & 6.6 \\ \hline

wav2vec2.0 Large  & LL-60k & Transformer & 2.9 & 5.8 \\ 
HuBERT Large & LL-60k & Transformer & 2.9 & 5.4 \\ 
WavLM Large & MIX-94k & Transformer & 2.9  & 5.1 \\ 
\midrule

\multicolumn{5}{c}{\textbf{\textit{10-hour labeled}}} \\ \hline
wav2vec 2.0   & LS-960 & None  & 11.1 & 17.6 \\ 
WavLM Base & LS-960 & None & 9.8 & 16.0 \\ 
WavLM Base+ & MIX-94k & None & 9.0 & 14.7 \\ \hline

DeCoAR 2.0   & LS-960 & 4-gram & 5.4 & 13.3  \\
DiscreteBERT   & LS-960 & 4-gram & 5.9 & 14.1  \\
wav2vec 2.0   & LS-960 & 4-gram  & 4.3 & 9.5 \\
HuBERT Base  & LS-960 & 4-gram & 4.3 & 9.4 \\ 
WavLM Base & LS-960 &  4-gram &  4.3  & 9.2   \\ 
WavLM Base+ & MIX-94k & 4-gram & 4.2  &  8.8 \\ 

wav2vec 2.0 Large   & LL-60k & 4-gram & 3.0 & 5.8 \\
WavLM Large & MIX-94k & 4-gram & 2.9  & 5.5 \\ \hline

wav2vec 2.0 Large  & LL-60k & Transformer & 2.6 & 4.9 \\
HuBERT Large   & LL-60k & Transformer & 2.4 & 4.6 \\  
WavLM Large & MIX-94k & Transformer & 2.4  &  4.6 \\
\midrule

\multicolumn{5}{c}{\textbf{\textit{100-hour labeled}}} \\ \hline
wav2vec 2.0 Base   & LS-960 & None & 6.1 & 13.3 \\ 
WavLM Base  & LS-960 & None & 5.7 & 12.0  \\ 
WavLM Base+ & MIX-94k & None & 4.6 & 10.1 \\ \hline

DeCoAR 2.0   & LS-960 & 4-gram & 5.0 & 12.1  \\
DiscreteBERT  & LS-960 & 4-gram &  4.5 & 12.1 \\
wav2vec 2.0 Base   & LS-960& 4-gram  & 3.4 & 8.0 \\
HuBERT Base   & LS-960 & 4-gram  & 3.4 & 8.1\\
WavLM Base & LS-960 &  4-gram &  3.4  &  7.7  \\ 
WavLM Base+ & MIX-94k & 4-gram & 2.9  &  6.8 \\ 

wav2vec 2.0 Large   & LL-60k & 4-gram & 2.3 & 4.6 \\ 
WavLM Large & MIX-94k & 4-gram & 2.3 & 4.6 \\ \hline

wav2vec 2.0 Large   & LL-60k & Transformer & 2.0 & 4.0 \\
HuBERT Large   & LL-60k & Transformer & 2.1 & 3.9 \\  
WavLM Large & MIX-94k & Transformer & 2.1 &  4.0 \\
 \hline
\midrule
\end{tabular}}
    
\end{center}
\end{table}
\raggedbottom

\begin{table}[t]
\begin{center}

\caption{WER on LibriSpeech when using all 960 hours of labeled data.}
\label{asr_960_result}

\resizebox{\columnwidth}{!}{
\begin{tabular}{lcccc}
\toprule
Model & Unlabled Data &  LM &   test-clean & test-other \\ \hline
\midrule
\multicolumn{5}{c}{\textbf{\textit{Supervised}}} \\ \hline
CTC Transf \cite{synnaeve2019end}  & -& CLM+Transf. & 2.5 & 5.5 \\
S2S Transf. \cite{synnaeve2019end}  &  - & CLM+Transf. & 2.3  & 5.2 \\
Transf. Transducer \cite{zhang2020transformer} & - & Transf. & 2.0  & 4.6 \\ 
ContextNet \cite{contextnet}  & - & LSTM & 1.9  & 4.1  \\
Conformer Transducer \cite{conformer} & - & LSTM & 1.9  & 3.9 \\ \hline
\multicolumn{5}{c}{\textbf{\textit{Pre-training}}} \\ \hline
wav2vec 2.0 Large & LL-60k & Transformer & 1.8 & 3.3  \\
HuBERT Large &  LL-60k & Transformer & 1.9  &  3.3 \\ 
WavLM Large & MIX-94k & Transformer & 1.8  & 3.2 \\ \hline
\midrule
        \end{tabular}}
\end{center}
\end{table}
\raggedbottom

\subsection{Speech Recognition }
\subsubsection{Problem Formulation}
Given the input speech signal $\mathbf{x} = (x_1, ..., x_T)$, the goal of speech recognition is to generate the corresponding transcription $\mathbf{y} = (y_1,...,y_L)$, where $T$ and $L$ are the lengths of the speech and transcription, respectively. 

\subsubsection{Datasets}
We use LibriSpeech for our ASR experiments. 
For the fine-tuning, we consider four different partitions: 960 hours of transcribed LibriSpeech \cite{librispeech}, the train-clean-100 subset (100 hours labeled data), as well as the Libri-Light \cite{librilight} limited resource training subsets originally extracted from LibriSpeech, including train-10h (10 hours labeled data) and train-1h (1 hour labeled data).  We follow the evaluation protocol of Libri-Light for these splits and evaluate on the standard LibriSpeech  test-clean/other sets.

\subsubsection{Implementation Details}
The pre-trained models are fine-tuned for speech recognition by adding a randomly initialized linear projection layer on top of the Transformer encoder.  Models are optimized based on a CTC loss \cite{ctc} where we have 29 tokens for character targets plus a word boundary token. We apply a modified version of SpecAugment \cite{specaug} by masking time-steps and channels: we randomly select the starting positions with a predetermined probability and replace a span of ten subsequent time-steps with a mask embedding; different spans may overlap and we use the same masked time step embedding as the one used for pre-training. We also mask channels by choosing a number of channels as starting indices and then expanding to the subsequent 64 channels. Spans may overlap and the selected spans are set to zeros.

During fine-tuning, the convolutional encoder is always fixed and we freeze the Transformer encoder for the first 10k steps.  We optimize with Adam and a tri-stage rate schedule where the learning rate is warmed up for the first 10\% of the updates, held constant for the next 40\%, and then linearly decayed for the remainder.  The Base and Base+ models are fine-tuned on 8 GPUs with a batch size equivalent to 200 seconds of audio for each GPU. The Large model is fine-tuned on 24 GPUs with a batch size equivalent to 80 seconds of audio for each GPU.  We also use LayerDrop \cite{fan2019reducing,huang2016deep} at a rate of 0.05 for Base/Base+ and 0.1 for LARGE. The summary of the fine-tuning hyperparameter settings used for different labeled data setups can be found in Appendix \ref{ssec:hyper_downstream}.

For evaluation, we use wav2letter++ \cite{wav2letter++} beam search decoder with language model (LM) fused decoding as :
\begin{equation}
    {\rm log}p_{CTC}(\mathbf{y}|\mathbf{x}) + w_1 {\rm log} p_{LM}(\mathbf{y}) + w_2 |\mathbf{y}|
\end{equation}
where $w_1$ is the language model weight and $w_2$ is the word insertion weight. We consider a 4-gram model and a Transformer model, which are identical to \cite{wav2vec2}. The evaluation hyperparameters are also based on \cite{wav2vec2}.

\subsubsection{Results}
Table \ref{asr_main_result} presents the results for the low-resource setup, where the pre-trained models are fine-tuned on the 1 hour, 10 hours or 100 hours of labeled data. We compare our method with several competitive self-supervised approaches in the literature, including DeCoAR 2.0 \cite{decoar2}, DiscreteBERT \cite{vq_wav2vec_ft}, wav2vec 2.0 \cite{wav2vec2} and HuBERT \cite{hsu2021hubert}. Without LM fusion, the WavLM Base model outperforms wav2vec 2.0 by a large margin for all fine-tuning splits, indicating the superiority of our model architecture. Its performance matches or outperforms wav2vec 2.0 and HuBERT with LM. WavLM Base+ improves WavLM Base, especially on the test-other set,  indicating increasing the out-of-domain unlabeled data also works for ASR.  For the Large model, the observation is consistent that our method achieves comparable or better performance than the baselines.  Table \ref{asr_960_result} reports results on the full 960 hours of LibriSpeech data. Overall, the pre-training methods can outperform all supervised models and our model is on par with the two best pre-training results in this setting.

\section{Conclusion}
We present WavLM, a large-scale pre-trained model with 94k hour audio as inputs, to solve full stack speech processing tasks. WavLM extends the HuBERT framework to masked speech prediction and denoising modeling, enabling the pre-trained models to perform well on both ASR and non-ASR tasks. WavLM updates state-of-the-art results on the SUPERB, as well as the representative testsets of speaker verification, speech separation, and speaker diarization.  In contrast to previous SSL models, WavLM is not only effective for the ASR task but also has the potential to become the next-generation backbone network for speaker-related tasks.

In the future, we would like to scale up the model size to increase the model capability, as previous work has shown the benefits of more parameters \cite{zhang2021bigssl}. Meanwhile, the model compression technique is also worth trying due to the time constraint and limited test time resources in real scenarios. It is also a promising direction to jointly learn text and speech representation in a self-supervised pre-training framework \cite{ao2021speecht5}, as the huge amount of text data might increase the capability of speech content modeling.

{\appendices
\section{Hyperparamters for pre-training}
\label{ssec:hyper_pretrain}

Table~\ref{hypers} shows the hyperparameters used for pre-training our WavLM Base, Base+, and Large model, which are adapted from the previous work \cite{hsu2021hubert}.

\begin{table}[h]
\centering
\caption{Hyperparamters for pre-training WavLM models \label{hypers}. The unit in batch size computing is second. We use 32 V100 GPUs for base model training, and 64 V100 GPUs for large model.  }
\resizebox{\columnwidth}{!}{
\begin{tabular}{l|ccccc}
\toprule
Model    & pre-train data & update steps & learning rate & warmup steps & batch size\\ \hline
WavLM Base &  960h  & 400k  & 5e-4 & 32k & 350s \\ 
WavLM Base+ &  94kh  & 1.2M  & 5e-4 & 96k &  350s\\ 
WavLM Large & 94kh & 700k & 1.5e-3 & 32k &  720s \\ 
\bottomrule
\end{tabular}}
\end{table}
\raggedbottom

\section*{Settings of downstream tasks}
\label{ssec:set_downstream}
For the universal representation evaluation, we use the same settings for all the SUPERB tasks in accordance with the SUPERB policies \cite{superb}.

As for the four additional downstream tasks, including speaker verification, speaker diarization, speech separation, and speech recognition, the implementations are shown in Table~\ref{downstream}, following the previous works \cite{desplanques2020ecapa,kinoshita2021advances,chen2020continuous2,hsu2021hubert}.

\begin{table}[t]
\centering
\caption{Different settings of the downstream tasks. In the speaker diarization task, it should be noted that the CALLHOME dataset is used for domain adaptation. \label{downstream}}
\resizebox{\columnwidth}{!}{
\begin{tabular}{l|lll}
\toprule
                 Task    & dataset &  training data duration & downstream model \\ \hline
                 Speaker Verification & VoxCeleb & 2300h& ECAPA-TDNN\\
                 Speaker Diarization & CALLHOME & 8.7h&Transformer \\
                 Speech Separation  & LibriCSS & 219h & Conformer\\
                 Speech Recognition & LibriSpeech & 1h/10h/100h/960h& Linear\\
\bottomrule

\end{tabular}
}

\end{table}
\raggedbottom

\section*{Hyperparamters for fine-tuning}
\label{ssec:hyper_downstream}

As for the universal representation evaluation, Table~\ref{table:superb_hyper} shows the hyperparameters of the learning rate and batch size for fine-tuning our WavLM models in the SUPERB downstream tasks. 
For the QbE task, which is evaluated by dynamic time warping without fine-tuning, we find that the best results for all the three WavLM models are always from the representations of the last layer.
All the other hyperparameters of each downstream task are exactly the same as the official implementation of SUPERB\footnote{\url{https://github.com/s3prl/s3prl}}.

As for the speech recognition task fine-tuning, Table~\ref{asr_ft_hypers} summarizes the hyperparameters used for different labeled data setups.

\begin{table}[h]
\centering
\caption{Hyperparamenters of fine-tuning WavLM models in SUPERB downstream tasks. The batch size of Speech Translation task denotes the number of tokens in each training batch.
}\label{table:superb_hyper}
\resizebox{\columnwidth}{!}{
\begin{tabular}{l|cc|cc|cc}
\toprule
\multirow{2}{*}{Task}    & \multicolumn{2}{c|}{WavLM Base} & \multicolumn{2}{c|}{WavLM Base+} & \multicolumn{2}{c}{WavLM Large} \\ 
    & learning rate & batch size   & learning rate & batch size  & learning rate & batch size \\ \hline
Speaker Identification &  2e-1   & 512 & 1e-1 &  512  & 5e-2 & 512  \\ 
Automatic Speaker Verification &  5e-5 & 512 & 5e-5  & 512  & 5e-5 &  512 \\ 
Speaker Diarization & 2e-3 & 256  & 5e-4 &  256 & 5e-3 &  256  \\ 
Phoneme Recognition & 5e-4 & 128  & 5e-4 & 128  & 2e-4 & 128 \\ 
Automatic Speech Recognition & 5e-4 &  128 & 5e-4 & 128  &  1e-4 & 128   \\
Out-of-domain Automatic Speech Recognition & 1e-4 & 16  & 1e-4 & 16  &  1e-4 & 16 \\
Keyword Spotting & 1e-5 &  512 & 1e-5 &  512 &  1e-5  &  512 \\
Speech Translation & 1e-3 & 80k  & 1e-3 & 80k  & 1e-3  &  160k \\
Intent Classification & 5e-5 & 128  & 2e-5 & 128  &  5e-4 & 128   \\
Slot Filling & 2e-4 & 128  & 2e-4 & 128  &  1e-4  &  128 \\
Emotion Recognition & 1e-4 & 32  & 1e-4 & 32  &  1e-5  &  32 \\
Speech Enhancement & 5e-4 & 64  & 5e-4 & 64  &  5e-4  &  64 \\
 Speech Separation & 5e-4 & 64  & 1e-3 & 64  &  5e-4  &  64 \\
Voice Conversion & 1e-4 & 6  & 1e-4 & 6  &  1e-4 & 6 \\
\bottomrule
\end{tabular}}
\end{table}
\raggedbottom

\begin{table}[!h]
\begin{center}
\caption{Hyperparamenters of fine-tuning WavLM models in speech recognition task.
\label{asr_ft_hypers}}
\resizebox{\columnwidth}{!}{
\begin{tabular}{l|cccc}
\toprule
Setup                     & updates &  learning rate & timestep mask prob. & channel mask prob. \\ \hline
1 hour (Base/Base+)  & 13k     &  5e-5          &   0.065 & 0.004          \\
10 hour (Base/Base+) & 25k     &  2e-5          &  0.075  & 0.008   \\
100 hour (Base/Base+) & 80k   &  3e-5  &  0.065  & 0.008  \\
1 hour (Large)     &   13k     &    3e-4  &  0.075 & 0.004   \\
10 hour (Large)     &   20k     &    1e-4   &   0.075 & 0.004   \\
100 hour (Large)     &   80k     &   3e-5   &  0.005 & 0.008  \\
960 hour (Large)     &  320k      &    3e-5   &   0.005 & 0.004   \\

\bottomrule
\end{tabular}}

\end{center}

\end{table}
\raggedbottom
}

\bibliography{refs}
\bibliographystyle{IEEEtran}

\vfill

\end{document}